\documentclass[acmlarge]{acmart}
\AtBeginDocument{%
  \providecommand\BibTeX{{%
    \normalfont B\kern-0.5em{\scshape i\kern-0.25em b}\kern-0.8em\TeX}}}

\setcopyright{acmcopyright}
\copyrightyear{2018}
\acmYear{2018}
\acmDOI{XXXXXXX.XXXXXXX}

\acmJournal{POMACS}
\acmVolume{37}
\acmNumber{4}
\acmArticle{111}
\acmMonth{8}

\usepackage{graphicx}
\usepackage{xcolor}
\usepackage{colortbl}
\usepackage{rotating}
\usepackage{booktabs}
\usepackage{multirow}
\usepackage{tikz}
\usepackage{wasysym}
\usepackage{pgfplots}
\pgfplotsset{compat=newest}
\usepackage{fontawesome5}
\usepackage{subcaption}
\usepackage{tabularx}
\usepackage{array}

\definecolor{casecolor}{HTML}{E2E9F1}
\definecolor{examplerow}{HTML}{F3F3F3}
\definecolor{errorred}{HTML}{D32F2F}

\begin{document}

\title{TextReasoningBench: Does Reasoning Really Improve Text Classification in Large Language Models?}

\author{Xinyu Guo}
\affiliation{%
  \institution{School of Computer Science and Technology, Tianjin University}
  \country{China}
}
\email{patrick2319674@gmail.com}

\author{Yazhou Zhang}
\authornote{Corresponding author.}
\affiliation{%
  \institution{School of Computer Science and Technology, Tianjin University}
  \country{China}}
\email{yzhou\_zhang@tju.edu.cn}

\author{Jing Qin}
\affiliation{%
  \institution{School of Nursing, The Hong Kong Polytechnic University}
  \country{HongKong}}
\email{harry.qin@polyu.edu.hk}

\renewcommand{\shortauthors}{Guo and Zhang, et al.}

\begin{abstract}
Eliciting explicit, step-by-step reasoning traces from large language models (LLMs) has emerged as a dominant paradigm for enhancing model capabilities. Although such reasoning strategies were originally designed for problems requiring explicit multi-step reasoning, they have increasingly been applied to a broad range of NLP tasks.
This expansion implicitly assumes that deliberative reasoning uniformly benefits heterogeneous tasks.
However, whether such reasoning mechanisms truly benefit classification tasks remains largely underexplored, especially considering their substantial token and time costs.
To fill this gap, we introduce \textit{TextReasoningBench}, a systematic benchmark designed to evaluate the effectiveness and efficiency of reasoning strategies for text classification with LLMs. We compare seven reasoning strategies, namely IO, CoT, SC-CoT, ToT, GoT, BoC, and long-CoT across ten LLMs on five text classification datasets.
Beyond traditional metrics such as accuracy and macro-F1, we introduce two cost-aware evaluation metrics that quantify the performance gain per reasoning token and the efficiency of performance improvement relative to token cost growth.
Experimental results reveal three notable findings:
(1) \textbf{Reasoning does not universally improve classification performance}: while moderate strategies such as CoT and SC-CoT yield consistent but limited gains (typically +1\% to +3\% on big models), more complex methods (e.g., ToT and GoT) often fail to outperform simpler baselines and can even degrade performance, especially on small models; 
(2) \textbf{Reasoning is often inefficient}: many reasoning strategies increase token consumption by 10$\times$ to 100$\times$ (e.g., SC-CoT and ToT) while providing only marginal performance improvements; 
(3) \textbf{Longer reasoning is not necessarily better}: performance exhibits a non-monotonic relationship with reasoning length, where moderate-length reasoning is beneficial but excessive reasoning leads to overthinking. 
These findings challenge the common assumption that explicit reasoning is inherently beneficial for NLP tasks.
\end{abstract}

\ccsdesc[500]{Computing methodologies~Natural language generation}

\keywords{text classification, sentiment analysis, reasoning strategies, large language models}

\maketitle

\section{Introduction}
Chain-of-thought (CoT) reasoning~\cite{wei2022chain}, inspired by the dual-process theory that distinguishes between fast, intuitive \textit{System-1} thinking and slow, deliberate \textit{System-2} reasoning, has inaugurated the era of large reasoning models (LRMs).
By eliciting intermediate reasoning steps, CoT and its descendants such as self-consistency (SC)~\cite{wang2022self}, tree-of-thought (ToT)~\cite{yao2023tree}, graph-of-thought (GoT)~\cite{besta2024graph}, and long-CoT~\cite{jaech2024openai} have achieved remarkable success on tasks that require explicit multi-step inference, including mathematical problem solving~\cite{ahn2024large}, logical reasoning~\cite{parmar2024logicbench}, and symbolic reasoning~\cite{he2026large}. 
The striking gains observed on these reasoning-intensive benchmarks have created a powerful precedent: explicit reasoning has rapidly become the default paradigm, applied broadly across a wide spectrum of NLP tasks under the implicit assumption that deliberate, step-by-step thinking is universally beneficial.
\begin{table*}[t]
\centering
\caption{Comparison of TextReasoningBench with existing reasoning benchmarks.}
\label{tab:compare}
\resizebox{\textwidth}{!}{%
\begin{tabular}{lcccccc}
\toprule
\textbf{Benchmark} & \textbf{Task Domain} & \textbf{Subjective Tasks} & \textbf{Analytical Tasks} & \textbf{\# Datasets} & \textbf{\# Reasoning Strategies} & \textbf{Cost-aware Evaluation} \\
\midrule
Mathador~\cite{kurtic2024mathador}       & Mathematical Reasoning & \faTimes & \faCheckSquare & 3 & 1 & \faTimes \\
MR-GSM8K~\cite{zeng2023mr}               & Mathematical Reasoning & \faTimes & \faCheckSquare & 1 & 2 & \faTimes \\
LogicBench~\cite{parmar2024logicbench}   & Logical Inference      & \faTimes & \faCheckSquare & 2 & 2 & \faTimes \\
CRUXEval~\cite{gu2024cruxeval}           & Code Reasoning         & \faTimes & \faCheckSquare & 1 & 1 & \faTimes \\
SarcasmBench~\cite{zhang2025sarcasmbench}& Sarcasm Detection      & \faCheckSquare & \faTimes & 6 & 3 & \faTimes \\
CARP~\cite{sun-etal-2023-text}           & Text Classification    & \faCheckSquare & \faTimes & 5 & 2 & \faTimes \\
\midrule
\textbf{TextReasoningBench} & \textbf{Text Classification} & \textbf{\faCheckSquare} & \textbf{\faCheckSquare} & \textbf{5} & \textbf{7} & \textbf{\faCheckSquare} \\
\bottomrule
\end{tabular}
}
\end{table*}

This assumption, however, deserves careful scrutiny. Unlike mathematical proofs or logical puzzles, many NLP tasks do not inherently require formal multi-step inference. Text classification, in particular, relies primarily on semantic understanding, distributional patterns, and contextual interpretation rather than explicit logical deduction. 
This mismatch becomes even more pronounced in subjective language understanding tasks such as sentiment analysis and sarcasm detection, where predictions hinge on contextual interpretation, pragmatic cues, and implicit cognitive understanding, none of which naturally decompose into explicit reasoning chains.  If the structural inductive bias of reasoning strategies is misaligned with the cognitive demands of these tasks, then reasoning may provide little benefit.
Compounding this concern, reasoning strategies typically generate extended chains of intermediate thoughts, incurring substantial increases in both token consumption and inference latency. This tension raises a fundamental question: \textit{does reasoning actually improve text classification?}

Despite the growing popularity of reasoning strategies, this question remains largely unexplored. 
Existing reasoning benchmarks, such as Mathador~\cite{kurtic2024mathador}, MR-GSM8K~\cite{zeng2023mr}, LogicBench~\cite{parmar2024logicbench}, and Cruxeval~\cite{gu2024cruxeval}, primarily focus on mathematical reasoning, commonsense inference, and code reasoning. 
In contrast, one of the most fundamental NLP problems, text classification, has received surprisingly limited attention in this context. 
The few related studies either evaluate only a small subset of reasoning strategies (e.g., CARP~\cite{sun-etal-2023-text}) or concentrate on a single task domain, such as sarcasm detection in SarcasmBench~\cite{zhang2025sarcasmbench}, as shown in Tab.~\ref{tab:compare}. 
Crucially, no existing work systematically examines the efficiency cost of reasoning strategies across classification settings, leaving open the question of under what conditions reasoning genuinely benefits text classification and at what computational price.

To fill this gap, we introduce \textit{TextReasoningBench}, a comprehensive evaluation framework for assessing the effectiveness and efficiency of reasoning strategies in text classification. We conduct experiments across both objective tasks (i.e., AGNews and TREC-QC) and subjective tasks (i.e., SST-2, SemEval 2018, and iSarcasmEval), covering four small LLMs (SLMs) and five big models (BMs). We compare seven representative reasoning methods, including standard input–output (IO) prompting, CoT, SC-CoT, ToT, GoT, bagging of cues (BoC), and long-CoT. All experiments are conducted in a zero-shot setting, with each configuration repeated five times to account for model stochasticity. Beyond conventional accuracy and macro-F1 metrics, we further propose two efficiency measures: \textit{absolute efficiency} (F1 per token) and \textit{marginal efficiency} (incremental F1 gain per additional token), jointly quantifying whether whether additional reasoning cost yields proportional gains.

Our experiments yield three key findings: 
(1) reasoning does not consistently improve classification performance, while CoT and SC-CoT provide modest gains (typically +1\% to +3\% on big models), more complex strategies such as ToT and GoT often fail to outperform IO and can even degrade performance; 
(2) reasoning is often highly inefficient, where many methods increase token consumption by 10$\times$ to 100$\times$ (e.g., SC-CoT and ToT) while yielding only marginal performance improvements, resulting in substantially lower absolute efficiency and frequently negative marginal efficiency; 
(3) longer reasoning is not inherently better. Performance exhibits a non-monotonic relationship with reasoning length, where moderate-length reasoning is beneficial but excessive reasoning leads to diminishing returns, instability (e.g., ToT showing variance up to 3.88), and overthinking, particularly for smaller models. 
These results challenge the prevailing assumption that more complex or longer reasoning universally improves performance across NLP tasks.
In summary, our main contributions are as follows:
\begin{itemize}
    \item We present \textbf{TextReasonBench}, the first comprehensive benchmark evaluating seven reasoning strategies across ten LLMs and five 
    text classification datasets.
    
    \item We propose two complementary efficiency metrics, \textbf{PCR} and 
    \textbf{ME}, to jointly assess the performance and cost of reasoning strategies.
    
    \item We provide extensive empirical evidence that reasoning strategies do not 
    universally benefit text classification, and that task subjectivity is a 
    critical yet previously overlooked moderator of reasoning effectiveness.
    
    \item We release all code, prompts, and evaluation results to facilitate 
    reproducibility and future research.\footnote{\url{https://github.com/codingNoob2319/TextReasoningBench-2026}}
\end{itemize}

\section{Related Work}
\subsection{LLM Reasoning}
Inspired by the human ability to reason step by step, CoT prompting was proposed to encourage language models to generate explicit intermediate reasoning steps. Wei et al.~\cite{wei2022chain} formally introduced CoT prompting and demonstrated its effectiveness on arithmetic and multi-step reasoning benchmarks. However, CoT relies heavily on carefully designed, high-quality demonstrations, which often require substantial manual effort. To mitigate this limitation, Auto-CoT~\cite{zhang2022automatic} was proposed to automatically construct reasoning exemplars, reducing dependence on human-crafted prompts.

Building on single-path reasoning, subsequent research shifted from single-path generation to multi-path exploration. Wang et al.~\cite{wang2022self} introduced self-consistency, which samples multiple independent reasoning trajectories and selects the final answer via majority voting. A few extensions further refined this paradigm, including adaptive self-consistency (ASC)~\cite{wan2024dynamic} and difficulty-adaptive self-consistency~\cite{wang2025make}, which dynamically adjust the number of sampled reasoning paths, aiming to balance performance and computational cost.

Beyond linear and multi-path reasoning, researchers have explored non-linear reasoning 
structures. Yao et al.~\cite{yao2023tree} proposed ToT modeling reasoning as a tree-structured search that enables exploration and backtracking across 
multiple branches. Besta et al.~\cite{besta2024graph} generalized this to 
GoT, organizing reasoning units as a directed graph to support richer information aggregation..

Most recently, the field has witnessed the rise of long-CoT and slow-thinking models. LRMs such as OpenAI o1/o3 and DeepSeek-R1~\cite{guo2025deepseek} internalize extended reasoning processes through reinforcement learning, enabling the generation of thousands of intermediate steps. These models achieve state-of-the-art (SoTA) performance on mathematical 
competitions, code generation, and symbolic reasoning. However, their inference cost is substantially higher than that of standard CoT or ToT methods and often suffer from overthinking.

Notably, virtually all of the above methods were proposed and evaluated on 
\textit{system-2} tasks, implicitly treating reasoning as a universally beneficial 
mechanism. The effectiveness of reasoning strategies on text classification tasks remains systematically unexplored, a gap this work directly addresses.

\subsection{LLMs Based Text Classification}
LLM-based text classification research generally follows two directions: model adaptation through fine-tuning and reasoning-based prompting.

The first line of work focuses on adapting LLMs to classification tasks via instruction tuning or architectural refinement. Wang et al.~\cite{wang2024improving} investigated how LLMs could be leveraged to produce higher-quality text representations for downstream classification. Building on this, Zhang et al.~\cite{zhang2025pushing} proposed an adaptive boosting framework that constructs multiple base learners through iterative reweighting and fine-tuning, subsequently ensembling them into a specialized classification model. Beyond objective classification, subjective classification has also received considerable attention. Konstantinidis et al.~\cite{konstantinidis2024finllama} fine-tuned Llama 2-7B to classify sentiment valence and quantify its intensity in financial text. Zhang et al.~\cite{zhang2025dialoguellm} presented DialogueLLM, which reformulates sentiment classification as a text generation problem and fine-tunes foundation models with contextual and emotional knowledge, yielding significant improvements across dialogue benchmarks. Cheng et al.~\cite{cheng2024emotion} further extended this direction by integrating multimodal inputs through emotion-specific encoders and instruction tuning, enhancing both emotion recognition and reasoning capabilities. Collectively, these approaches improve classification by modifying the internal representations or parameters of LLMs, leaving the reasoning structure at inference time unchanged.

The second line of research examines whether introducing explicit reasoning steps at inference time can improve classification performance without modifying model parameters. Ma et al.~\cite{ma2024chain} proposed Chain-of-Stance (CoS), which decomposes stance detection into a sequence of intermediate stance-related judgments, enabling more structured and interpretable predictions. Sun et al.~\cite{sun-etal-2023-text} introduced CARP, a prompting strategy that first directs LLMs to identify surface-level clues—such as keywords, tone, and semantic relations—before performing diagnostic reasoning to arrive at the final prediction. Wang et al.~\cite{wang2021novel} designed the Multi-Label Reasoner (ML-Reasoner), a reasoning-based framework for multi-label classification that employs a binary classifier for joint label prediction and incorporates an iterative reasoning mechanism to capture inter-label dependencies.

More recently, reasoning-based prompting has been extended to subjective classification tasks such as sentiment analysis and sarcasm detection, where the relationship between surface form and underlying meaning is particularly indirect. Fei et al.~\cite{fei2023reasoning} proposed a three-hop reasoning framework that sequentially infers implicit aspects, opinions, and sentiment polarity. Lai et al.~\cite{lai2025rvisa} introduced a two-stage reasoning-and-verification framework to enhance the reliability of implicit sentiment classification. In the context of sarcasm detection, Yao et al.~\cite{yao2025sarcasm} proposed SarcasmCue, a prompting framework comprising four complementary sub-methods, chain of contradiction (CoC), graph of cues (GoC), bagging of cues (BoC), and tensor of cues (ToC) that encourage LLMs to detect sarcasm through both sequential and non-sequential reasoning strategies. Song et al.~\cite{song2025emotion} further presented the first long-CoT framework for sentiment classification, which dynamically adjusts reasoning depth according to task complexity across multiple datasets.

Despite these advances, existing studies share a common limitation: they typically evaluate only a narrow set of reasoning strategies or confine their analysis to a single task domain. Crucially, none provides a systematic comparison of diverse reasoning paradigms across both objective and subjective classification tasks, nor do they account for the token and computational costs associated with different reasoning strategies. TextReasoningBench is designed to directly address these gaps.

\section{Methodology}
\subsection{Task Formulation}
We consider a standard text classification setting. 
Let $\mathcal{X}$ denote the space of input texts and 
$\mathcal{Y} = \{1, \dots, C\}$ denote a finite label set with $C$ classes.
Given an input text $x = (x_1, x_2, \dots, x_n) \in \mathcal{X}$, 
where each $x_i$ is a token, the goal is to predict a label $y \in \mathcal{Y}$.

We denote a pre-trained LLM as $p_\theta$ with parameters $\theta$. 
The model defines a conditional probability distribution over output sequences $p_\theta(y \mid \text{prompt}(x))$.
In the zero-shot setting, the model parameters $\theta$ remain fixed, and task instructions are provided through prompting. 
The prompt function $\text{prompt}(x)$ wraps the input text $x$ with task-specific instructions. 
The predicted output $\hat{y}$ is obtained by
\begin{equation}
\hat{y} = \arg\max_{y \in \mathcal{Y}} p_\theta(y \mid \text{prompt}(x))
\end{equation}

For reasoning-based prompting, the model may additionally generate intermediate reasoning steps $z = (z_1, \dots, z_m)$ before producing the final prediction. The joint generation process can be expressed as:
\begin{equation}
(z, y) \sim p_\theta(z, y \mid \text{prompt}(x)),
\end{equation}
where $z$ denotes the reasoning trace and $y$ is the final predicted label.
All reasoning strategies studied in this work differ only in how $z$ is constructed or aggregated.
The LLM parameters $\theta$ remain unchanged throughout evaluation.

\subsection{Reasoning Strategies}\label{sec:reasoning}
\begin{figure}[t]
    \centering
    \includegraphics[width=0.7\linewidth]{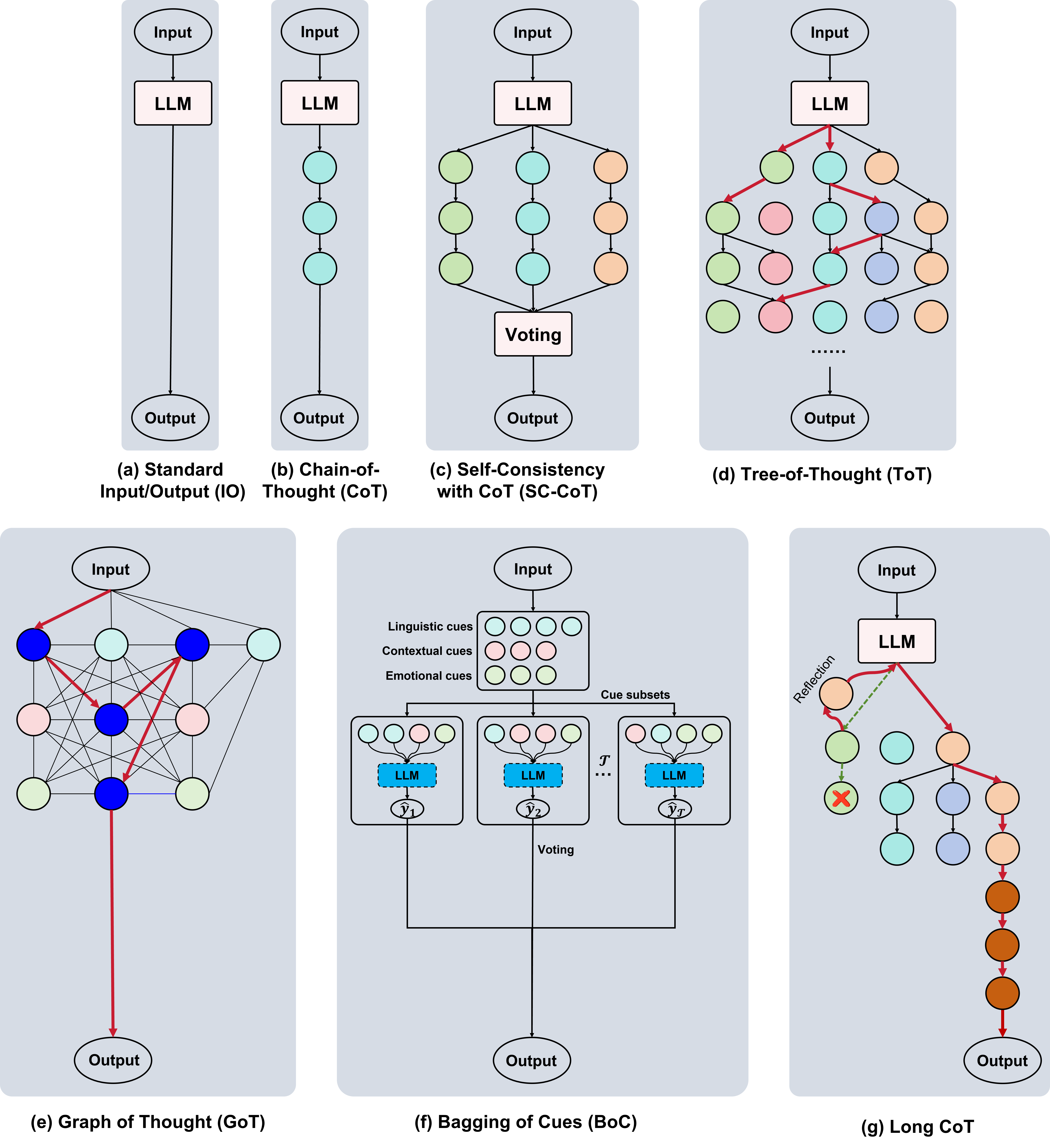}
    \caption{Seven reasoning methods.}
    \label{fig:models}
\end{figure}

We select seven typical reasoning methods covering linear and non-linear paradigms, as shown in Fig.~\ref{fig:models}.

\textbf{(1) Input–Output Prompting (IO).}
IO prompting is the most straightforward approach. 
Given an input text $x$, the model directly generates a prediction without producing explicit intermediate reasoning steps:
\begin{equation}
\hat{y} = \arg\max_{y \in \mathcal{Y}} p_\theta(y \mid \text{prompt}(x))
\end{equation}

In this setting, the mapping from $x$ to $y$ is treated as a single conditional generation process. 
No reasoning trace $z$ is explicitly constructed. 
IO therefore represents the minimal inference cost configuration and provides a reference point for evaluating the added value of reasoning strategies.

\textbf{(2) CoT Prompting.}
CoT prompting introduces an explicit reasoning trace $z = (z_1, \dots, z_m)$ that decomposes the decision process into intermediate steps $(z, y) \sim p_\theta(z, y \mid \text{prompt}(x))$.
Each reasoning step is generated sequentially:
\begin{equation}
z_i \sim p_\theta(z_i \mid x, z_{<i}), 
\quad 
y \sim p_\theta(y \mid x, z)
\end{equation}

CoT expands the output space by embedding classification within a longer generative trajectory. 
This effectively transforms a direct mapping $x \to y$ into a conditional sequence modeling problem $x \to z \to y$.
Structurally, CoT enforces a linear reasoning topology. 
While it may reduce decision variance by encouraging more deliberate processing, it also increases token cost proportionally to reasoning length $m$ and remains confined to a single reasoning path.

\textbf{(3) SC-CoT.}
Self-consistency addresses the variance of single-chain CoT by sampling multiple independent reasoning trajectories $(z^{(k)}, y^{(k)}) \sim p_\theta(z, y \mid \text{prompt}(x)), 
\quad k = 1, \dots, K$. The final prediction is selected via majority voting:
\begin{equation}
\hat{y} = \arg\max_{y \in \mathcal{Y}} \sum_{k=1}^{K} \mathbf{1}(y^{(k)} = y)
\end{equation}

SC-CoT can be interpreted as Monte Carlo marginalization over latent reasoning paths. 
Instead of committing to a single trajectory, it approximates the expectation over multiple plausible chains.
This approach often improves robustness by mitigating stochastic decoding errors. 
However, inference cost scales linearly with the number of samples $K$, making it significantly more expensive than standard CoT.

\textbf{(4) ToT.}
ToT extends linear reasoning into a structured search process. 
Instead of generating a single chain, ToT explores multiple candidate reasoning states organized as a tree. 
At reasoning depth $t$, multiple branches are expanded $z_{t+1}^{(b)} \sim p_\theta(z \mid x, z_{\leq t})$, where $b$ indexes branching decisions.
This branching mechanism enables local exploration and backtracking. 
The search may be guided by heuristic scoring functions that evaluate partial reasoning states. 
The final prediction is derived from the highest-scoring branch.

Compared with SC-CoT, which samples independent full chains, ToT performs structured exploration within each reasoning trajectory. 
However, its search space grows exponentially with branching factor and depth, leading to substantially increased computational complexity.

\textbf{(5) GoT.} GoT generalizes ToT by allowing reasoning states to form an arbitrary directed graph $\mathcal{G} = (\mathcal{V}, \mathcal{E})$. 
Nodes $\mathcal{V}$ represent reasoning units, and edges $\mathcal{E}$ encode dependencies between them.

Unlike tree structures, GoT allows merging, revisiting, and recombining reasoning states across branches. Information can propagate non-sequentially, enabling richer structural interactions. The final prediction is obtained by aggregating information across the reasoning graph:
\begin{equation}
\hat{y} = g(\mathcal{G}(x))
\end{equation}

While GoT increases expressive flexibility and may capture complex reasoning dependencies, it further expands the search space and introduces additional computational overhead.

\textbf{(6) BoC.}
Bagging-of-cue is proposed based on the set theory, which emphasizes diversity over structure. 
Instead of enforcing sequential reasoning or structured search, BoC generates multiple independent reasoning fragments: $z^{(k)} \sim p_\theta(z \mid \text{prompt}(x))$, where $k = 1, \dots, K$.

Each fragment yields a tentative prediction, and aggregation (e.g., voting or scoring) produces the final output.
BoC can be viewed as a lightweight ensemble over partial reasoning samples. 
Compared with SC-CoT, it does not require each reasoning instance to form a coherent chain, thereby reducing structural constraints while maintaining diversity. 
Its computational cost scales with $K$ but remains simpler than structured search approaches.

\textbf{(7) Long-CoT.}
Long-CoT significantly extends the reasoning depth by generating prolonged reasoning sequences:
\begin{equation}
(z_1, z_2, \dots, z_T, y) \sim p_\theta(\cdot \mid \text{prompt}(x))
\end{equation}
where $T \gg m$ in standard CoT.

Long-CoT often incorporates self-reflection, intermediate evaluation, or dynamic continuation mechanisms. It increases the effective reasoning horizon, allowing deeper deliberation before producing the final label.

However, the substantially increased token consumption may introduce diminishing returns. 
Excessively long reasoning chains can amplify noise accumulation and lead to overthinking, making efficiency analysis particularly important in classification settings.

\subsection{Evaluation Protocol}
To ensure fair and reproducible comparisons across reasoning strategies, 
all experiments are conducted under a zero-shot setting. 
Model parameters remain frozen throughout evaluation, and no fine-tuning, training data, or external knowledge sources are used. 
Any performance differences therefore arise solely from differences in reasoning structure rather than parameter adaptation.

To eliminate prompt-induced variance, we adopt a unified prompt template shared by all reasoning strategies. 
Each method uses the same task instruction, input format, and output format. 
The base prompt template is defined as follows:
\begin{quote}
\textbf{Instruction:} Please classify the following text into one of the predefined categories. The output must be exactly one label from the candidate set.  

\textbf{Input:} Text: \{x\}. Label set: \{y$_1$, y$_2$, \dots, y$_C$\}  

\textbf{Output:} Answer: <label>
\end{quote}

For reasoning-based strategies (e.g., CoT, SC-CoT, long-CoT), the following instruction is appended before prediction: \textit{Let's think step by step}. Each method further follows its dedicated reasoning structure as described in Sec.~\ref{sec:reasoning}. Regardless of the reasoning structure, the final prediction must strictly follow the format 
``Answer: <label>'' to ensure consistent parsing and evaluation.

Due to stochastic decoding in LLMs, each configuration (model $\times$ task $\times$ reasoning strategy) is executed five independent times. 
We report the mean and standard deviation across the five runs to assess both performance and stability. 
To ensure comparability across all configurations, a unified set of decoding 
hyperparameters is applied: temperature $= 0.9$, top-$p = 0.95$, frequency 
penalty $= 0$, and presence penalty $= 0$. We adopt temperature $0.9$ uniformly 
across all strategies to ensure that observed performance 
differences are attributable solely to reasoning structure rather than decoding 
regime. For standard strategies, the maximum generation length is fixed at 512 tokens per call. 
For Long-CoT, we allow a larger generation budget (up to 2,048 tokens) to accommodate longer reasoning chains, while avoiding excessively large token allocations that would confound efficiency comparisons. The total token consumption per instance is reported separately as part of our efficiency analysis in Sec.~\ref{sec:efficiency}.

\subsection{Performance Metrics}
In addition to conventional classification metrics such as \textit{accuracy} and \textit{macro-F1}, 
we introduce two efficiency-oriented measures to quantify whether the additional computational cost 
introduced by reasoning yields proportional performance gains. 
These metrics explicitly account for the trade-off between effectiveness and inference cost.

\textbf{Token Cost.}
For each reasoning strategy $s$, we define its token cost $T_s$ as the total number of tokens consumed over the test set, including both input tokens and generated tokens. 
To ensure comparability across strategies, all methods are evaluated under identical prompts and datasets.

\paragraph{Absolute Efficiency.}
Absolute Efficiency (AE) measures the performance achieved per unit of token consumption. 
Let $\text{F1}_s$ denote the Macro-F1 score of strategy $s$, and $T_s$ its total token cost. 
We define:
\begin{equation}
\text{AE}_s = \frac{\text{F1}_s}{T_s}
\end{equation}


AE captures the overall cost-effectiveness of a reasoning strategy. 
A higher AE indicates that the model achieves stronger predictive performance relative to its computational cost. 
Notably, AE penalizes excessively long reasoning traces that yield only marginal improvements.

\paragraph{Marginal Efficiency.}
While AE reflects overall efficiency, it does not quantify whether additional reasoning beyond a baseline is worthwhile. 
To measure incremental benefit, we introduce Marginal Efficiency (ME). Let $\text{F1}_0$ and $T_0$ denote the Macro-F1 score and token cost of the IO baseline, respectively. 
For a reasoning strategy $s$, define:
\begin{equation}
\Delta \text{F1}_s = \text{F1}_s - \text{F1}_0, \quad
\Delta T_s = T_s - T_0
\end{equation}

We define:
\begin{equation}
    \text{ME}_s = \frac{\Delta\text{F1}_s}{\Delta T_s} \times 10^3
\end{equation}
ME measures the incremental macro-F1 gain per additional token expended 
relative to IO. The sign and magnitude of ME jointly characterize four 
qualitatively distinct regimes, as summarized in 
Table~\ref{tab:me_interpretation}.

\begin{table}[t]
\small
\centering
\caption{Interpretation of ME across different regimes.}\label{tab:me_interpretation}
\begin{tabular}{ccccp{7.5cm}}
\toprule
 \textbf{Notation} & $\Delta\text{F1}_s$ & $\Delta T_s$ & $\text{ME}_s$ & \textbf{Interpretation} \\
\midrule
 $\bigstar $ & $> 0$ & $> 0$ & $> 0$ & Reasoning improves performance at a cost; higher ME indicates better return on investment \\
 $\clubsuit  $ & $> 0$ & $< 0$ & $< 0$ & Reasoning improves performance \textit{and} reduces cost; strategy dominates IO \\
 $ \spadesuit $ & $< 0$ & $> 0$ & $< 0$ & Reasoning degrades performance while increasing cost; strategy is harmful \\
 $\maltese $ & $< 0$ & $< 0$ & $> 0$ & Reasoning degrades performance but reduces cost; ME is misleadingly positive \\
\bottomrule
\end{tabular}
\end{table}

To avoid misinterpretation in the fourth regime (performance loss with cost 
reduction), ME should always be interpreted jointly with the sign of 
$\Delta\text{F1}_s$. We therefore report $\Delta\text{F1}_s$, $\Delta T_s$, 
and $\text{ME}_s$ together in our efficiency analysis. When $\Delta T_s 
\approx 0$ (i.e., a strategy incurs negligible additional token cost relative to 
IO), ME is undefined and we report AE only.

\paragraph{Discussion.}
Together, AE and ME provide complementary perspectives on reasoning efficiency. 
AE evaluates overall cost-effectiveness across strategies, while ME focuses on the marginal utility of additional reasoning beyond a baseline. 
This dual perspective enables a more principled analysis of whether increasingly complex reasoning strategies are justified in text classification.

\section{Experiments}
\subsection{Experimental Setup}
We evaluate seven popular reasoning strategies (IO, CoT, SC-CoT, ToT, GoT, BoC, and long-CoT) across both objective and subjective text classification tasks. 
We report the mean and standard deviation of Accuracy, Macro-F1, and Weighted-F1. 
In addition, we compute AE and ME to analyze the trade-off between predictive performance and inference cost.

\begin{table}[htbp] 
    \centering
    \small
    \caption{Detailed statistics of the evaluated datasets.}
    \label{tab:datasets}
    \begin{tabular}{lcccc}
        \toprule
        \textbf{Dataset} & \textbf{Task Type} & \textbf{Classes} & \textbf{Avg. Length} & \textbf{\#Eval Samples} \\
        \midrule
        AG News (Subset) & Objective & 4 & 37.3 & 7,600 \\
        TREC-QC & Objective & 6 & 7.5 & 500 \\
        SST-2 & Subjective & 2 & 19.5 & 872 \\
        SemEval 2018 Task 3 & Subjective & 2 & 14.6 & 784 \\
        iSarcasmEval & Subjective & 2 & 16.4 & 1,400 \\
        \bottomrule
    \end{tabular}
\end{table}
\subsection{Datasets}
We select five widely used benchmark datasets 
covering both objective and subjective text classification tasks, as shown in Tab.~\ref{tab:datasets}.

\paragraph{Objective Classification Tasks.} 
\textbf{AGNews}~\cite{zhang2015character} is a four-class topic classification dataset derived from the AG's Corpus of News Articles. It contains 120,000 training samples 
and 7,600 test samples, categorized into \textit{World}, \textit{Sports}, 
\textit{Business}, and \textit{Sci/Tech}. 
The task primarily involves factual topic categorization based on explicit semantic cues. 
Since topic boundaries are largely content-driven and relatively stable, 
AGNews represents a objective classification task.

\textbf{TREC-QC} (Question Classification)~\cite{li2002experimental} is a six-class benchmark consisting of 5,452 training and 500 test questions. 
Questions are categorized into semantic types such as \textit{Person}, 
\textit{Location}, \textit{Entity}, \textit{Description}, \textit{Number}, 
and \textit{Abbreviation}. Average length of each sentence is 10, vocabulary size of 8700.
The task requires identifying the expected answer type of a question, which depends primarily on syntactic and lexical patterns. 

\paragraph{Subjective Classification Tasks.}

\textbf{SST-2} (Stanford Sentiment Treebank)~\cite{socher-etal-2013-recursive} 
is a binary sentiment classification dataset containing 67,349 training samples. 
Since the official test labels are withheld, following common practice, we 
evaluate on the validation split comprising 872 samples. Each sentence is labeled 
as either \textit{positive} or \textit{negative}. Unlike topic classification, 
sentiment analysis often requires interpreting implicit polarity, contextual 
nuances, and affective expressions, making it a representative benchmark for 
subjective language understanding.

\textbf{SemEval 2018 Task 3 (Irony Detection)}~\cite{van2018semeval} consists of 3,834 training and 784 test tweets annotated for irony. The dataset focuses on detecting ironic expressions in social media, 
where surface meaning frequently contradicts intended meaning. 

\textbf{iSarcasmEval}~\cite{farha2022semeval} is a benchmark for sarcasm detection based on author-annotated tweets, where sarcasm labels are provided by the original authors of the tweets rather than third-party annotators, ensuring high label reliability. The task is formulated as binary classification: sarcastic or non-sarcastic.

\subsection{Comparative Models}
We select nine SoTA LLMs spanning a wide range of model scales. 

\paragraph{Small LLMs (SLMs):}
\begin{itemize}
    \item \textbf{GPT-4o-mini}\footnote{https://developers.openai.com/api/docs/models/gpt-4o-mini} is a lightweight variant of the GPT-4o series, designed for cost-efficient inference while retaining strong language understanding capabilities.
    \item \textbf{Qwen-3-8B}\footnote{https://huggingface.co/Qwen/Qwen3-8B} is an 8-billion 
parameter model from Alibaba's Qwen3 series, supporting both thinking and non-thinking inference modes.
    \item \textbf{Llama-3.1-8B}\footnote{https://huggingface.co/meta-llama/Llama-3.1-8B} is an 
open-source model from Meta's Llama 3.1 family, instruction-tuned for general-purpose language tasks.
    \item \textbf{Gemma-3-4B}\footnote{https://huggingface.co/google/gemma-3-4b-it} is a compact model from Google's Gemma 3 family, optimized for efficient deployment with competitive performance.
\end{itemize}

\paragraph{Big Models (BMs):}
\begin{itemize}
    \item \textbf{GPT-5.2}\footnote{https://developers.openai.com/api/docs/models/gpt-5} represents the current strongest model with advanced reasoning and instruction-following capabilities. It is widely recognized for strong performance across complex reasoning benchmarks.
    \item \textbf{Gemini-2.5-Flash}\footnote{https://docs.cloud.google.com/vertex-ai/generative-ai/docs/models/gemini/2-5-flash?hl=zh-cn} is Google DeepMind's SoTA multimodal model with strong performance across a broad range of language benchmarks. 
    \item \textbf{Qwen-3-Max}\footnote{https://qwen.ai/home} is the largest publicly accessible variant in the Qwen-3 series, designed for high-performance reasoning and long-context understanding.
    \item \textbf{DeepSeek-V3.2}\footnote{https://huggingface.co/deepseek-ai/DeepSeek-V3.2} is a high-capacity mixture-of-experts model from DeepSeek, demonstrating competitive performance with frontier proprietary systems at significantly lower inference cost.
    \item \textbf{Kimi-K2}\footnote{https://huggingface.co/moonshotai/Kimi-K2.5} is a SoTA model developed by Moonshot AI, with particular strengths in long-context understanding and instruction following.
\end{itemize}

\begin{table}[t]
\centering
\small
\caption{Model versions and release dates.}
\label{tab:model_versions}
\begin{tabular}{lll}
\toprule
\textbf{Model} & \textbf{Version / Checkpoint} & \textbf{Release Date} \\
\midrule
GPT-4o-mini & gpt-4o-mini-2024-07-18 & 2024-07-18 \\
Qwen-3-8B & Qwen/Qwen3-8B-Instruct & 2025-03-14 \\
Llama-3.1-8B & meta-llama/Llama-3.1-8B-Instruct & 2024-07-23 \\
Gemma-3-4B & google/gemma-3-4b-it & 2025-03-10 \\
GPT-5.2 & gpt-5 (snapshot: 2025-12-11) & 2025-12-11 \\
Gemini-2.5-Flash & gemini-2.5-flash & 2025-06-17 \\
Qwen-3-Max & qwen3-max-2025-09-23 & 2025-09-23 \\
DeepSeek-V3.2 & deepseek-ai/DeepSeek-V3.2 & 2025-12-01 \\
Kimi-K2 & kimi-k2-0905-preview & 2025-09-05 \\
\bottomrule
\end{tabular}
\end{table}

\paragraph{Model Access and Versioning.}
All models are accessed via their official APIs using publicly available instruction-tuned versions. 
Open-weight models are deployed using their official checkpoints without additional fine-tuning, as shown in Table~\ref{tab:model_versions}. All evaluations are conducted between December 2025 and February 2026.
\begin{table*}[t]
\centering
\caption{Performance of different reasoning methods on \textbf{SLMs}. \textcolor{red}{\textbf{Red}} indicates the overall best result in each column, while \textcolor{blue}{\textbf{blue}} indicates the best result for each individual model. Metrics are presented in percentages as mean$_{\pm std}$.}
\label{tab:small_models}
\resizebox{\textwidth}{!}{%
\begin{tabular}{llcccccccccc}
\toprule
\multirow{2}{*}{\textbf{Model}} & \multirow{2}{*}{\textbf{Method}} & \multicolumn{2}{c}{\textbf{AG News}} & \multicolumn{2}{c}{\textbf{TREC-QC}} & \multicolumn{2}{c}{\textbf{SST-2}} & \multicolumn{2}{c}{\textbf{SemEval-2018}} & \multicolumn{2}{c}{\textbf{iSarcasmEval}} \\
\cmidrule(lr){3-4} \cmidrule(lr){5-6} \cmidrule(lr){7-8} \cmidrule(lr){9-10} \cmidrule(lr){11-12}
 &  & Acc. & Ma-F1 & Acc. & Ma-F1 & Acc. & Ma-F1 & Acc. & Ma-F1 & Acc. & Ma-F1 \\
\midrule
\multirow{6}{*}{GPT-4o-Mini} & IO & 83.13$_{{\pm 0.34}}$ & 82.75$_{{\pm 0.35}}$ & \textcolor{blue}{\textbf{86.27}}$_{{\pm 0.62}}$ & \textcolor{red}{\textbf{85.35}}$_{{\pm 0.87}}$ & 92.81$_{{\pm 0.11}}$ & 92.84$_{{\pm 0.10}}$ & 67.13$_{{\pm 0.26}}$ & 66.68$_{{\pm 0.28}}$ & 55.50$_{{\pm 0.41}}$ & 51.67$_{{\pm 0.30}}$ \\
 & CoT & \textcolor{red}{\textbf{86.93}}$_{{\pm 0.66}}$ & \textcolor{red}{\textbf{86.73}}$_{{\pm 0.71}}$ & 84.73$_{{\pm 0.66}}$ & 84.02$_{{\pm 0.37}}$ & 93.31$_{{\pm 0.05}}$ & 93.31$_{{\pm 0.05}}$ & 66.50$_{{\pm 0.75}}$ & 65.93$_{{\pm 0.84}}$ & 55.93$_{{\pm 0.32}}$ & 52.00$_{{\pm 0.24}}$ \\
 & SC-CoT & 86.70$_{{\pm 0.30}}$ & 86.55$_{{\pm 0.36}}$ & 83.00$_{{\pm 0.60}}$ & 82.87$_{{\pm 0.26}}$ & \textcolor{red}{\textbf{93.81}}$_{{\pm 0.23}}$ & \textcolor{blue}{\textbf{93.80}}$_{{\pm 0.23}}$ & 71.36$_{{\pm 0.57}}$ & 71.18$_{{\pm 0.60}}$ & 63.32$_{{\pm 0.46}}$ & 57.49$_{{\pm 0.34}}$ \\
 & ToT & 49.20$_{{\pm 2.80}}$ & 47.95$_{{\pm 3.46}}$ & 54.90$_{{\pm 0.70}}$ & 53.50$_{{\pm 0.61}}$ & 82.22$_{{\pm 0.34}}$ & 82.02$_{{\pm 0.28}}$ & \textcolor{blue}{\textbf{75.32}}$_{{\pm 0.06}}$ & \textcolor{blue}{\textbf{75.48}}$_{{\pm 0.09}}$ & \textcolor{blue}{\textbf{68.46}}$_{{\pm 0.32}}$ & \textcolor{blue}{\textbf{61.23}}$_{{\pm 0.40}}$ \\
 & GoT & 82.50$_{{\pm 0.10}}$ & 82.07$_{{\pm 0.13}}$ & 81.60$_{{\pm 0.40}}$ & 77.39$_{{\pm 0.57}}$ & 90.77$_{{\pm 0.17}}$ & 90.74$_{{\pm 0.17}}$ & 71.49$_{{\pm 0.45}}$ & 71.32$_{{\pm 0.47}}$ & 62.11$_{{\pm 0.25}}$ & 56.88$_{{\pm 0.10}}$ \\
 & BoC & 83.70$_{{\pm 0.30}}$ & 83.43$_{{\pm 0.33}}$ & 83.80$_{{\pm 0.00}}$ & 82.78$_{{\pm 0.25}}$ & 93.41$_{{\pm 0.29}}$ & 93.40$_{{\pm 0.29}}$ & 68.30$_{{\pm 0.32}}$ & 67.91$_{{\pm 0.35}}$ & 56.32$_{{\pm 0.11}}$ & 52.26$_{{\pm 0.17}}$ \\
\midrule
\multirow{6}{*}{Gemma-3-4B} & IO & 76.40$_{{\pm 0.33}}$ & 74.28$_{{\pm 0.34}}$ & 72.53$_{{\pm 0.25}}$ & 67.41$_{{\pm 0.28}}$ & 91.06$_{{\pm 0.09}}$ & 91.03$_{{\pm 0.09}}$ & 46.73$_{{\pm 0.06}}$ & 40.39$_{{\pm 0.09}}$ & 28.55$_{{\pm 0.07}}$ & 28.55$_{{\pm 0.07}}$ \\
 & CoT & 79.60$_{{\pm 0.59}}$ & 77.93$_{{\pm 0.66}}$ & 69.73$_{{\pm 1.11}}$ & 61.47$_{{\pm 0.29}}$ & \textcolor{blue}{\textbf{91.09}}$_{{\pm 0.46}}$ & \textcolor{blue}{\textbf{91.13}}$_{{\pm 0.50}}$ & \textcolor{blue}{\textbf{61.27}}$_{{\pm 0.85}}$ & \textcolor{blue}{\textbf{60.07}}$_{{\pm 0.97}}$ & \textcolor{blue}{\textbf{50.12}}$_{{\pm 0.62}}$ & \textcolor{blue}{\textbf{47.30}}$_{{\pm 0.58}}$ \\
 & SC-CoT & \textcolor{blue}{\textbf{80.70}}$_{{\pm 0.10}}$ & \textcolor{blue}{\textbf{79.61}}$_{{\pm 0.22}}$ & \textcolor{blue}{\textbf{74.20}}$_{{\pm 0.00}}$ & \textcolor{blue}{\textbf{70.54}}$_{{\pm 0.39}}$ & 90.42$_{{\pm 0.52}}$ & 90.40$_{{\pm 0.52}}$ & 56.06$_{{\pm 0.32}}$ & 53.68$_{{\pm 0.38}}$ & 39.14$_{{\pm 0.14}}$ & 38.33$_{{\pm 0.11}}$ \\
 & ToT & 40.50$_{{\pm 2.70}}$ & 38.88$_{{\pm 3.08}}$ & 32.10$_{{\pm 0.10}}$ & 32.37$_{{\pm 0.62}}$ & 81.88$_{{\pm 0.23}}$ & 82.47$_{{\pm 0.24}}$ & 40.43$_{{\pm 0.13}}$ & 35.06$_{{\pm 0.08}}$ & 21.79$_{{\pm 0.29}}$ & 22.68$_{{\pm 0.19}}$ \\
 & GoT & 68.70$_{{\pm 0.50}}$ & 67.97$_{{\pm 0.78}}$ & 65.70$_{{\pm 0.10}}$ & 61.46$_{{\pm 1.03}}$ & 73.91$_{{\pm 0.06}}$ & 72.36$_{{\pm 0.08}}$ & 42.60$_{{\pm 0.38}}$ & 34.54$_{{\pm 0.70}}$ & 21.86$_{{\pm 0.50}}$ & 21.54$_{{\pm 0.56}}$ \\
 & BoC & 77.20$_{{\pm 0.20}}$ & 75.03$_{{\pm 0.17}}$ & 65.80$_{{\pm 0.00}}$ & 65.01$_{{\pm 1.73}}$ & 90.88$_{{\pm 0.40}}$ & 90.89$_{{\pm 0.43}}$ & 45.79$_{{\pm 0.13}}$ & 38.92$_{{\pm 0.20}}$ & 25.61$_{{\pm 0.32}}$ & 25.55$_{{\pm 0.33}}$ \\
\midrule
\multirow{6}{*}{Llama-3.1-8B} & IO & 81.00$_{{\pm 0.43}}$ & 80.86$_{{\pm 0.47}}$ & 82.20$_{{\pm 1.07}}$ & 82.88$_{{\pm 0.84}}$ & 90.94$_{{\pm 0.32}}$ & 91.24$_{{\pm 0.32}}$ & \textcolor{blue}{\textbf{72.02}}$_{{\pm 1.38}}$ & \textcolor{blue}{\textbf{70.19}}$_{{\pm 1.32}}$ & \textcolor{blue}{\textbf{77.50}}$_{{\pm 0.36}}$ & \textcolor{blue}{\textbf{62.14}}$_{{\pm 0.58}}$ \\
 & CoT & \textcolor{blue}{\textbf{81.93}}$_{{\pm 0.19}}$ & \textcolor{blue}{\textbf{81.51}}$_{{\pm 0.22}}$ & 76.40$_{{\pm 1.56}}$ & 71.53$_{{\pm 2.46}}$ & 90.60$_{{\pm 0.32}}$ & 90.85$_{{\pm 0.36}}$ & 61.95$_{{\pm 1.36}}$ & 61.97$_{{\pm 1.47}}$ & 56.17$_{{\pm 0.55}}$ & 52.21$_{{\pm 0.32}}$ \\
 & SC-CoT & 81.40$_{{\pm 0.60}}$ & 80.56$_{{\pm 0.71}}$ & \textcolor{blue}{\textbf{85.10}}$_{{\pm 0.50}}$ & \textcolor{blue}{\textbf{82.91}}$_{{\pm 1.70}}$ & \textcolor{blue}{\textbf{92.60}}$_{{\pm 0.40}}$ & \textcolor{blue}{\textbf{92.60}}$_{{\pm 0.40}}$ & 66.33$_{{\pm 0.38}}$ & 66.04$_{{\pm 0.43}}$ & 60.04$_{{\pm 0.82}}$ & 55.13$_{{\pm 0.64}}$ \\
 & ToT & 71.70$_{{\pm 1.70}}$ & 72.57$_{{\pm 1.29}}$ & 68.00$_{{\pm 0.40}}$ & 70.18$_{{\pm 1.13}}$ & 83.77$_{{\pm 0.86}}$ & 84.03$_{{\pm 0.89}}$ & 61.99$_{{\pm 1.28}}$ & 61.83$_{{\pm 1.45}}$ & 54.39$_{{\pm 0.11}}$ & 50.57$_{{\pm 0.33}}$ \\
 & GoT & 81.90$_{{\pm 0.50}}$ & 81.46$_{{\pm 0.47}}$ & 78.80$_{{\pm 0.20}}$ & 77.05$_{{\pm 0.71}}$ & 91.46$_{{\pm 0.29}}$ & 91.52$_{{\pm 0.31}}$ & 66.58$_{{\pm 0.00}}$ & 66.29$_{{\pm 0.01}}$ & 60.79$_{{\pm 0.64}}$ & 55.31$_{{\pm 0.63}}$ \\
 & BoC & 77.50$_{{\pm 0.10}}$ & 76.89$_{{\pm 0.06}}$ & 82.80$_{{\pm 0.40}}$ & 82.08$_{{\pm 0.33}}$ & 91.28$_{{\pm 0.11}}$ & 91.32$_{{\pm 0.12}}$ & 51.91$_{{\pm 0.38}}$ & 48.19$_{{\pm 0.44}}$ & 33.43$_{{\pm 0.57}}$ & 33.28$_{{\pm 0.54}}$ \\
\midrule
\multirow{7}{*}{Qwen3-8B} & IO & 78.47$_{{\pm 0.09}}$ & 78.28$_{{\pm 0.10}}$ & 80.16$_{{\pm 0.32}}$ & 81.82$_{{\pm 0.31}}$ & 92.11$_{{\pm 0.26}}$ & 92.94$_{{\pm 0.19}}$ & 49.13$_{{\pm 0.37}}$ & 44.02$_{{\pm 0.55}}$ & 32.07$_{{\pm 0.20}}$ & 31.97$_{{\pm 0.19}}$ \\
 & CoT & 84.13$_{{\pm 0.50}}$ & 83.92$_{{\pm 0.57}}$ & 78.92$_{{\pm 0.59}}$ & 75.96$_{{\pm 1.74}}$ & 90.64$_{{\pm 0.33}}$ & 90.62$_{{\pm 0.33}}$ & 66.48$_{{\pm 0.41}}$ & 66.07$_{{\pm 0.50}}$ & 58.57$_{{\pm 0.38}}$ & 52.99$_{{\pm 0.36}}$ \\
 & SC-CoT & 82.40$_{{\pm 0.20}}$ & 82.01$_{{\pm 0.31}}$ & 83.30$_{{\pm 0.90}}$ & 79.70$_{{\pm 1.60}}$ & 91.93$_{{\pm 0.38}}$ & 91.93$_{{\pm 0.38}}$ & 75.13$_{{\pm 0.10}}$ & 75.11$_{{\pm 0.10}}$ & 69.46$_{{\pm 0.18}}$ & 61.19$_{{\pm 0.28}}$ \\
 & ToT & 63.00$_{{\pm 4.00}}$ & 63.79$_{{\pm 3.88}}$ & 56.80$_{{\pm 0.00}}$ & 54.07$_{{\pm 0.16}}$ & 82.65$_{{\pm 0.29}}$ & 82.76$_{{\pm 0.16}}$ & \textcolor{red}{\textbf{79.21}}$_{{\pm 1.56}}$ & \textcolor{red}{\textbf{78.74}}$_{{\pm 1.59}}$ & \textcolor{red}{\textbf{80.32}}$_{{\pm 0.04}}$ & \textcolor{red}{\textbf{67.74}}$_{{\pm 0.23}}$ \\
 & GoT & \textcolor{blue}{\textbf{84.30}}$_{{\pm 0.30}}$ & \textcolor{blue}{\textbf{83.94}}$_{{\pm 0.32}}$ & 83.07$_{{\pm 0.84}}$ & 80.07$_{{\pm 1.11}}$ & 91.36$_{{\pm 0.24}}$ & 91.34$_{{\pm 0.24}}$ & 72.83$_{{\pm 0.10}}$ & 72.77$_{{\pm 0.11}}$ & 65.11$_{{\pm 0.04}}$ & 58.33$_{{\pm 0.06}}$ \\
 & BoC & 77.10$_{{\pm 0.30}}$ & 76.28$_{{\pm 0.38}}$ & 78.30$_{{\pm 0.10}}$ & 77.56$_{{\pm 0.30}}$ & \textcolor{red}{\textbf{93.81}}$_{{\pm 0.31}}$ & \textcolor{red}{\textbf{93.81}}$_{{\pm 0.30}}$ & 63.20$_{{\pm 0.06}}$ & 62.33$_{{\pm 0.09}}$ & 54.44$_{{\pm 0.43}}$ & 50.26$_{{\pm 0.34}}$ \\
 & Long-CoT & 82.70$_{{\pm 0.70}}$ & 82.26$_{{\pm 0.71}}$ & \textcolor{red}{\textbf{87.10}}$_{{\pm 0.50}}$ & \textcolor{blue}{\textbf{83.56}}$_{{\pm 0.35}}$ & 92.49$_{{\pm 0.06}}$ & 92.53$_{{\pm 0.06}}$ & 73.02$_{{\pm 0.19}}$ & 72.96$_{{\pm 0.20}}$ & 68.18$_{{\pm 0.25}}$ & 60.37$_{{\pm 0.36}}$ \\
\bottomrule
\end{tabular}%
}
\end{table*}

\subsection{Main Results on Small Models}
Tab.~\ref{tab:small_models} presents the performance of different reasoning strategies on SLMs. 
We focus our analysis on Macro-F1 (Ma-F1), which provides a more robust evaluation under class imbalance.

\textbf{(1) Reasoning does not consistently improve classification performance.}
Across models and datasets, reasoning strategies do not uniformly outperform the IO baseline. 
In several cases, IO remains highly competitive or even superior. 
For example, on TREC-QC, GPT-4o-Mini achieves the best performance with IO (85.35 Ma-F1), outperforming all reasoning-based methods. 
Similarly, for Llama-3.1-8B, IO achieves the best results on SemEval-2018 (70.19) and iSarcasmEval (62.14), while CoT reduces performance by approximately 11.7\% and 16.0\%, respectively. 
These results suggest that, for many classification tasks, direct prediction without explicit reasoning is already sufficient and may even be preferable.

\textbf{(2) Lightweight reasoning provides moderate but task-dependent gains.}
Simple reasoning strategies such as CoT and SC-CoT can improve performance in certain settings, although the gains are generally modest and inconsistent. 
For instance, on AG News, CoT improves GPT-4o-Mini from 82.75 to 86.73 Ma-F1 (+4.81\%), and SC-CoT improves Gemma-3-4B from 74.28 to 79.61 (+7.18\%). 
On SST-2, SC-CoT consistently achieves strong results across models, but the improvements remain relatively small (typically within 1--2\%). 
These observations indicate that linear reasoning can provide useful inductive bias, but its effectiveness depends heavily on the task and model.

\textbf{(3) Complex reasoning is highly unstable and often harmful on objective tasks.}
More sophisticated reasoning paradigms, particularly ToT and GoT, exhibit significant instability and can severely degrade performance, especially on objective classification tasks. 
For example, ToT reduces GPT-4o-Mini performance on AG News from 82.75 to 47.95 Ma-F1 (-42.05\%) and on TREC-QC from 85.35 to 53.50 (-37.30\%). 
Similar degradation is observed for Gemma-3-4B and Qwen3-8B, where ToT leads to drops exceeding 30\% on TREC-QC. 
These results suggest that structured search introduces excessive noise and error propagation in tasks that do not require multi-step reasoning, leading to unstable decision boundaries.

\textbf{(4) Reasoning is more beneficial for subjective language understanding tasks.}
A clear contrast emerges between objective and subjective tasks. 
On datasets involving subjective interpretation, such as SemEval-2018 and iSarcasmEval, reasoning strategies—especially SC-CoT and ToT—often yield substantial improvements. 
For instance, on Qwen3-8B, ToT improves SemEval-2018 from 44.02 to 78.74 Ma-F1 (+78.87\%) and iSarcasmEval from 31.97 to 67.74 (+111.89\%). 
Similarly, on Gemma-3-4B, CoT improves iSarcasmEval from 28.55 to 47.30 (+65.67\%). 
These gains indicate that reasoning can help capture implicit semantics, pragmatic cues, and contextual nuances that are critical for subjective classification.

\textbf{(5) Long-CoT provides limited benefits for small models.}
Extending reasoning depth does not consistently improve performance. 
For Qwen3-8B, Long-CoT achieves only marginal gains on TREC-QC (83.56 vs.\ 81.82, +2.13\%) and does not outperform simpler methods on most other tasks. 
For example, it underperforms ToT on SemEval-2018 (72.96 vs.\ 78.74) and iSarcasmEval (60.37 vs.\ 67.74), and is inferior to BoC on SST-2 (92.53 vs.\ 93.81). 
These results suggest that increasing reasoning length alone is insufficient for improving classification, and may introduce redundancy and noise, particularly for small models with limited capacity.

\textbf{(6) Model sensitivity to reasoning varies significantly.}
Different models exhibit distinct responses to reasoning strategies. 
GPT-4o-Mini benefits moderately from CoT but suffers from ToT on objective tasks. 
Gemma-3-4B relies heavily on simple reasoning (e.g., CoT and SC-CoT), yet is highly vulnerable to complex reasoning structures. 
Llama-3.1-8B shows strong baseline performance with IO and limited gains from reasoning, while Qwen3-8B benefits substantially from reasoning on subjective tasks but not on objective ones. 
This highlights that the effectiveness of reasoning is jointly determined by task characteristics and model capacity.

\textbf{Summary.}
Overall, these results challenge the common assumption that more sophisticated reasoning universally improves classification performance. 
Instead, reasoning effectiveness is highly task-dependent, model-dependent, and complexity-sensitive. 
Simple or even non-reasoning strategies often achieve competitive performance on objective tasks, while more complex reasoning is only beneficial when it aligns with the underlying cognitive demands of the task.

\begin{table*}[t]
\centering
\caption{Performance of different reasoning methods on \textbf{BMs}. \textcolor{red}{\textbf{Red}} indicates the overall best result in each column, while \textcolor{blue}{\textbf{blue}} indicates the best result for each individual model. Metrics are presented in percentages as mean$_{\pm std}$.}
\label{tab:large_models}
\resizebox{\textwidth}{!}{%
\begin{tabular}{llcccccccccc}
\toprule
\multirow{2}{*}{\textbf{Model}} & \multirow{2}{*}{\textbf{Method}} & \multicolumn{2}{c}{\textbf{AG News}} & \multicolumn{2}{c}{\textbf{TREC-QC}} & \multicolumn{2}{c}{\textbf{SST-2}} & \multicolumn{2}{c}{\textbf{SemEval-2018}} & \multicolumn{2}{c}{\textbf{iSarcasmEval}} \\
\cmidrule(lr){3-4} \cmidrule(lr){5-6} \cmidrule(lr){7-8} \cmidrule(lr){9-10} \cmidrule(lr){11-12}
 &  & Acc. & Ma-F1 & Acc. & Ma-F1 & Acc. & Ma-F1 & Acc. & Ma-F1 & Acc. & Ma-F1 \\
\midrule
\multirow{7}{*}{DeepSeek-V3.2} & IO & 82.33$_{{\pm 0.09}}$ & 81.45$_{{\pm 0.08}}$ & 86.32$_{{\pm 0.27}}$ & 83.59$_{{\pm 0.30}}$ & 93.99$_{{\pm 0.09}}$ & 94.02$_{{\pm 0.08}}$ & 71.60$_{{\pm 0.16}}$ & 71.47$_{{\pm 0.17}}$ & 58.80$_{{\pm 0.21}}$ & 53.68$_{{\pm 0.17}}$ \\
 & CoT & 84.73$_{{\pm 0.66}}$ & 84.37$_{{\pm 0.69}}$ & 87.00$_{{\pm 0.46}}$ & 86.85$_{{\pm 0.74}}$ & 94.13$_{{\pm 0.27}}$ & 94.12$_{{\pm 0.27}}$ & 76.45$_{{\pm 0.32}}$ & 76.44$_{{\pm 0.32}}$ & 66.93$_{{\pm 0.33}}$ & 60.59$_{{\pm 0.26}}$ \\
 & SC-CoT & 85.40$_{{\pm 0.40}}$ & 85.05$_{{\pm 0.49}}$ & 88.00$_{{\pm 0.20}}$ & 87.83$_{{\pm 0.13}}$ & \textcolor{blue}{\textbf{95.36}}$_{{\pm 0.06}}$ & \textcolor{blue}{\textbf{95.35}}$_{{\pm 0.06}}$ & 72.00$_{{\pm 0.32}}$ & 71.85$_{{\pm 0.33}}$ & 64.43$_{{\pm 0.07}}$ & 58.71$_{{\pm 0.02}}$ \\
 & ToT & 64.20$_{{\pm 1.40}}$ & 65.35$_{{\pm 1.43}}$ & 73.40$_{{\pm 0.40}}$ & 68.37$_{{\pm 0.02}}$ & 83.83$_{{\pm 0.23}}$ & 83.68$_{{\pm 0.26}}$ & 74.36$_{{\pm 0.51}}$ & 74.16$_{{\pm 0.47}}$ & 69.36$_{{\pm 1.00}}$ & 60.84$_{{\pm 0.94}}$ \\
 & GoT & 81.00$_{{\pm 0.00}}$ & 80.21$_{{\pm 0.09}}$ & 87.13$_{{\pm 1.04}}$ & 85.29$_{{\pm 1.73}}$ & 93.31$_{{\pm 0.29}}$ & 93.30$_{{\pm 0.29}}$ & 75.38$_{{\pm 0.31}}$ & 75.38$_{{\pm 0.31}}$ & 68.36$_{{\pm 0.07}}$ & 62.12$_{{\pm 0.21}}$ \\
 & BoC & 84.40$_{{\pm 0.40}}$ & 83.95$_{{\pm 0.44}}$ & 88.40$_{{\pm 0.00}}$ & \textcolor{blue}{\textbf{87.84}}$_{{\pm 0.63}}$ & 94.72$_{{\pm 0.00}}$ & 94.72$_{{\pm 0.00}}$ & 75.38$_{{\pm 0.26}}$ & 75.35$_{{\pm 0.26}}$ & 65.89$_{{\pm 0.18}}$ & 59.73$_{{\pm 0.17}}$ \\
 & Long-CoT & \textcolor{blue}{\textbf{86.50}}$_{{\pm 0.30}}$ & \textcolor{blue}{\textbf{86.32}}$_{{\pm 0.34}}$ & \textcolor{blue}{\textbf{89.30}}$_{{\pm 0.30}}$ & 86.03$_{{\pm 1.27}}$ & 94.90$_{{\pm 0.17}}$ & 94.90$_{{\pm 0.17}}$ & \textcolor{blue}{\textbf{78.19}}$_{{\pm 1.15}}$ & \textcolor{blue}{\textbf{78.19}}$_{{\pm 1.15}}$ & \textcolor{blue}{\textbf{70.86}}$_{{\pm 0.57}}$ & \textcolor{blue}{\textbf{63.95}}$_{{\pm 0.51}}$ \\
\midrule
\multirow{7}{*}{GPT-5.2} & IO & 87.80$_{{\pm 0.59}}$ & \textcolor{blue}{\textbf{87.86}}$_{{\pm 0.49}}$ & 93.00$_{{\pm 0.57}}$ & 92.58$_{{\pm 0.75}}$ & 94.53$_{{\pm 0.05}}$ & 95.26$_{{\pm 0.09}}$ & 85.37$_{{\pm 0.49}}$ & 85.87$_{{\pm 0.40}}$ & 82.79$_{{\pm 0.20}}$ & 74.95$_{{\pm 0.30}}$ \\
 & CoT & 86.53$_{{\pm 0.50}}$ & 86.69$_{{\pm 0.35}}$ & 92.60$_{{\pm 0.43}}$ & 92.26$_{{\pm 0.62}}$ & 95.18$_{{\pm 0.25}}$ & 95.31$_{{\pm 0.19}}$ & 85.46$_{{\pm 0.45}}$ & 85.99$_{{\pm 0.42}}$ & 82.93$_{{\pm 0.06}}$ & 75.25$_{{\pm 0.20}}$ \\
 & SC-CoT & \textcolor{blue}{\textbf{88.00}}$_{{\pm 0.00}}$ & 87.81$_{{\pm 0.00}}$ & 93.80$_{{\pm 0.00}}$ & 92.83$_{{\pm 0.00}}$ & 95.53$_{{\pm 0.00}}$ & 95.53$_{{\pm 0.00}}$ & 86.48$_{{\pm 0.00}}$ & 86.53$_{{\pm 0.00}}$ & 83.71$_{{\pm 0.00}}$ & 75.87$_{{\pm 0.00}}$ \\
 & ToT & 87.40$_{{\pm 0.00}}$ & 87.21$_{{\pm 0.00}}$ & 93.60$_{{\pm 0.00}}$ & 92.72$_{{\pm 0.00}}$ & \textcolor{red}{\textbf{95.87}}$_{{\pm 0.00}}$ & \textcolor{red}{\textbf{95.93}}$_{{\pm 0.00}}$ & 87.63$_{{\pm 0.00}}$ & 87.74$_{{\pm 0.00}}$ & 84.43$_{{\pm 0.00}}$ & 76.25$_{{\pm 0.00}}$ \\
 & GoT & 87.40$_{{\pm 0.00}}$ & 87.15$_{{\pm 0.00}}$ & 92.60$_{{\pm 0.00}}$ & 92.50$_{{\pm 0.00}}$ & 95.64$_{{\pm 0.00}}$ & 95.64$_{{\pm 0.00}}$ & \textcolor{red}{\textbf{88.90}}$_{{\pm 0.00}}$ & \textcolor{red}{\textbf{88.78}}$_{{\pm 0.00}}$ & \textcolor{red}{\textbf{87.36}}$_{{\pm 0.00}}$ & 79.31$_{{\pm 0.00}}$ \\
 & BoC & 87.60$_{{\pm 0.60}}$ & 87.36$_{{\pm 0.67}}$ & 92.10$_{{\pm 0.30}}$ & 91.40$_{{\pm 0.26}}$ & 95.53$_{{\pm 0.11}}$ & 95.55$_{{\pm 0.09}}$ & 85.20$_{{\pm 0.26}}$ & 85.35$_{{\pm 0.28}}$ & 81.00$_{{\pm 0.29}}$ & 72.86$_{{\pm 0.34}}$ \\
 & Long-CoT & 87.20$_{{\pm 0.60}}$ & 87.71$_{{\pm 0.31}}$ & \textcolor{red}{\textbf{94.40}}$_{{\pm 0.20}}$ & \textcolor{red}{\textbf{94.07}}$_{{\pm 0.09}}$ & 95.64$_{{\pm 0.00}}$ & 95.83$_{{\pm 0.03}}$ & 86.93$_{{\pm 0.57}}$ & 87.81$_{{\pm 0.47}}$ & 86.50$_{{\pm 0.07}}$ & \textcolor{red}{\textbf{79.35}}$_{{\pm 0.02}}$ \\
\midrule
\multirow{7}{*}{Gemini-2.5-Flash} & IO & \textcolor{blue}{\textbf{85.60}}$_{{\pm 1.02}}$ & \textcolor{blue}{\textbf{85.31}}$_{{\pm 1.10}}$ & 88.13$_{{\pm 0.77}}$ & 85.84$_{{\pm 1.49}}$ & 94.42$_{{\pm 0.29}}$ & 94.42$_{{\pm 0.29}}$ & \textcolor{blue}{\textbf{79.34}}$_{{\pm 0.28}}$ & \textcolor{blue}{\textbf{79.33}}$_{{\pm 0.27}}$ & 74.90$_{{\pm 0.71}}$ & 67.00$_{{\pm 0.64}}$ \\
 & CoT & 85.00$_{{\pm 0.86}}$ & 84.68$_{{\pm 0.92}}$ & 87.60$_{{\pm 0.82}}$ & 86.14$_{{\pm 0.73}}$ & 94.34$_{{\pm 0.14}}$ & 94.34$_{{\pm 0.14}}$ & 74.79$_{{\pm 0.87}}$ & 74.77$_{{\pm 0.87}}$ & 72.32$_{{\pm 0.04}}$ & 64.94$_{{\pm 0.03}}$ \\
 & SC-CoT & 84.80$_{{\pm 0.00}}$ & 84.43$_{{\pm 0.00}}$ & 88.40$_{{\pm 0.60}}$ & 83.89$_{{\pm 0.53}}$ & 94.04$_{{\pm 0.00}}$ & 94.04$_{{\pm 0.00}}$ & 78.57$_{{\pm 0.00}}$ & 78.57$_{{\pm 0.00}}$ & \textcolor{blue}{\textbf{75.00}}$_{{\pm 0.00}}$ & \textcolor{blue}{\textbf{67.28}}$_{{\pm 0.00}}$ \\
 & ToT & 85.00$_{{\pm 0.00}}$ & 84.82$_{{\pm 0.00}}$ & 82.00$_{{\pm 0.00}}$ & 74.76$_{{\pm 0.00}}$ & 92.78$_{{\pm 0.00}}$ & 92.78$_{{\pm 0.00}}$ & 75.89$_{{\pm 0.00}}$ & 75.89$_{{\pm 0.00}}$ & 72.07$_{{\pm 0.00}}$ & 64.49$_{{\pm 0.00}}$ \\
 & GoT & 82.40$_{{\pm 0.00}}$ & 81.75$_{{\pm 0.00}}$ & 88.40$_{{\pm 0.00}}$ & 83.83$_{{\pm 0.00}}$ & 93.69$_{{\pm 0.00}}$ & 93.69$_{{\pm 0.00}}$ & 73.09$_{{\pm 0.00}}$ & 73.06$_{{\pm 0.00}}$ & 67.14$_{{\pm 0.00}}$ & 60.67$_{{\pm 0.00}}$ \\
 & BoC & 85.00$_{{\pm 0.60}}$ & 84.70$_{{\pm 0.58}}$ & \textcolor{blue}{\textbf{89.60}}$_{{\pm 0.00}}$ & \textcolor{blue}{\textbf{87.84}}$_{{\pm 0.33}}$ & \textcolor{blue}{\textbf{95.07}}$_{{\pm 0.00}}$ & \textcolor{blue}{\textbf{95.07}}$_{{\pm 0.00}}$ & 72.45$_{{\pm 0.00}}$ & 72.33$_{{\pm 0.00}}$ & 65.93$_{{\pm 0.00}}$ & 59.84$_{{\pm 0.00}}$ \\
 & Long-CoT & 85.20$_{{\pm 0.00}}$ & 84.94$_{{\pm 0.00}}$ & 89.20$_{{\pm 0.00}}$ & 84.53$_{{\pm 0.00}}$ & 94.61$_{{\pm 0.00}}$ & 94.61$_{{\pm 0.00}}$ & 77.68$_{{\pm 0.00}}$ & 77.68$_{{\pm 0.00}}$ & 73.86$_{{\pm 0.00}}$ & 66.33$_{{\pm 0.00}}$ \\
\midrule
\multirow{7}{*}{Kimi-K2} & IO & 84.93$_{{\pm 0.66}}$ & 84.82$_{{\pm 0.75}}$ & 87.88$_{{\pm 0.45}}$ & 88.16$_{{\pm 0.61}}$ & \textcolor{blue}{\textbf{94.68}}$_{{\pm 0.19}}$ & \textcolor{blue}{\textbf{94.72}}$_{{\pm 0.19}}$ & 71.73$_{{\pm 0.42}}$ & 71.61$_{{\pm 0.43}}$ & 61.36$_{{\pm 0.56}}$ & 56.41$_{{\pm 0.46}}$ \\
 & CoT & 86.80$_{{\pm 0.75}}$ & 86.92$_{{\pm 0.77}}$ & 90.72$_{{\pm 0.39}}$ & 88.65$_{{\pm 0.66}}$ & 93.72$_{{\pm 0.30}}$ & 93.71$_{{\pm 0.30}}$ & 70.92$_{{\pm 0.78}}$ & 70.79$_{{\pm 0.82}}$ & 62.34$_{{\pm 0.59}}$ & 57.15$_{{\pm 0.51}}$ \\
 & SC-CoT & \textcolor{blue}{\textbf{87.10}}$_{{\pm 0.10}}$ & \textcolor{blue}{\textbf{87.16}}$_{{\pm 0.11}}$ & 90.67$_{{\pm 0.57}}$ & 90.06$_{{\pm 0.74}}$ & 94.50$_{{\pm 0.23}}$ & 94.49$_{{\pm 0.23}}$ & 71.05$_{{\pm 0.38}}$ & 70.85$_{{\pm 0.40}}$ & 61.93$_{{\pm 0.07}}$ & 56.81$_{{\pm 0.02}}$ \\
 & ToT & 79.60$_{{\pm 0.40}}$ & 79.80$_{{\pm 0.46}}$ & 79.20$_{{\pm 2.00}}$ & 74.52$_{{\pm 1.70}}$ & 87.73$_{{\pm 0.46}}$ & 88.04$_{{\pm 0.41}}$ & 73.15$_{{\pm 0.96}}$ & 73.29$_{{\pm 0.99}}$ & 65.96$_{{\pm 0.46}}$ & 60.04$_{{\pm 0.29}}$ \\
 & GoT & 82.10$_{{\pm 1.30}}$ & 81.94$_{{\pm 1.35}}$ & 84.90$_{{\pm 0.90}}$ & 84.40$_{{\pm 1.74}}$ & 92.55$_{{\pm 0.11}}$ & 92.54$_{{\pm 0.11}}$ & 73.15$_{{\pm 0.19}}$ & 73.14$_{{\pm 0.20}}$ & 66.32$_{{\pm 0.18}}$ & 60.06$_{{\pm 0.14}}$ \\
 & BoC & 86.90$_{{\pm 0.50}}$ & 86.89$_{{\pm 0.46}}$ & \textcolor{blue}{\textbf{91.10}}$_{{\pm 0.10}}$ & \textcolor{blue}{\textbf{90.22}}$_{{\pm 0.58}}$ & 94.21$_{{\pm 0.29}}$ & 94.21$_{{\pm 0.29}}$ & 75.51$_{{\pm 0.38}}$ & 75.51$_{{\pm 0.38}}$ & 72.11$_{{\pm 0.25}}$ & 64.95$_{{\pm 0.30}}$ \\
 & Long-CoT & 86.90$_{{\pm 0.30}}$ & 86.95$_{{\pm 0.27}}$ & 90.00$_{{\pm 0.80}}$ & 87.63$_{{\pm 0.73}}$ & 93.92$_{{\pm 0.00}}$ & 93.92$_{{\pm 0.00}}$ & \textcolor{blue}{\textbf{80.10}}$_{{\pm 0.00}}$ & \textcolor{blue}{\textbf{80.06}}$_{{\pm 0.01}}$ & \textcolor{blue}{\textbf{77.18}}$_{{\pm 0.04}}$ & \textcolor{blue}{\textbf{68.93}}$_{{\pm 0.06}}$ \\
\midrule
\multirow{7}{*}{Qwen3-Max} & IO & 87.93$_{{\pm 0.19}}$ & 87.88$_{{\pm 0.19}}$ & 90.20$_{{\pm 0.13}}$ & 88.43$_{{\pm 0.35}}$ & 94.75$_{{\pm 0.05}}$ & 94.75$_{{\pm 0.05}}$ & 78.98$_{{\pm 0.22}}$ & 78.98$_{{\pm 0.22}}$ & 68.27$_{{\pm 0.12}}$ & 61.86$_{{\pm 0.12}}$ \\
 & CoT & 88.67$_{{\pm 0.25}}$ & 88.61$_{{\pm 0.26}}$ & 89.88$_{{\pm 0.30}}$ & 89.53$_{{\pm 0.73}}$ & 94.61$_{{\pm 0.19}}$ & 94.61$_{{\pm 0.19}}$ & 77.07$_{{\pm 0.38}}$ & 77.06$_{{\pm 0.38}}$ & 68.94$_{{\pm 0.40}}$ & 62.49$_{{\pm 0.45}}$ \\
 & SC-CoT & \textcolor{red}{\textbf{88.70}}$_{{\pm 0.30}}$ & \textcolor{red}{\textbf{88.64}}$_{{\pm 0.29}}$ & 89.80$_{{\pm 0.20}}$ & 89.59$_{{\pm 0.22}}$ & 95.24$_{{\pm 0.06}}$ & 95.24$_{{\pm 0.06}}$ & \textcolor{blue}{\textbf{80.55}}$_{{\pm 0.06}}$ & \textcolor{blue}{\textbf{80.51}}$_{{\pm 0.06}}$ & \textcolor{blue}{\textbf{74.96}}$_{{\pm 0.54}}$ & \textcolor{blue}{\textbf{67.43}}$_{{\pm 0.54}}$ \\
 & ToT & 76.80$_{{\pm 0.20}}$ & 77.07$_{{\pm 0.24}}$ & 84.90$_{{\pm 0.30}}$ & 79.09$_{{\pm 0.64}}$ & 90.60$_{{\pm 1.26}}$ & 90.59$_{{\pm 1.27}}$ & 73.72$_{{\pm 0.51}}$ & 73.28$_{{\pm 0.42}}$ & 74.04$_{{\pm 0.11}}$ & 65.40$_{{\pm 0.23}}$ \\
 & GoT & 87.90$_{{\pm 0.30}}$ & 87.79$_{{\pm 0.29}}$ & 89.73$_{{\pm 0.38}}$ & 89.53$_{{\pm 0.34}}$ & 93.92$_{{\pm 0.25}}$ & 93.92$_{{\pm 0.25}}$ & 79.00$_{{\pm 0.53}}$ & 78.99$_{{\pm 0.53}}$ & 70.93$_{{\pm 0.15}}$ & 63.85$_{{\pm 0.08}}$ \\
 & BoC & 87.80$_{{\pm 0.20}}$ & 87.65$_{{\pm 0.22}}$ & \textcolor{blue}{\textbf{91.00}}$_{{\pm 0.00}}$ & \textcolor{blue}{\textbf{89.95}}$_{{\pm 0.03}}$ & 94.32$_{{\pm 0.06}}$ & 94.32$_{{\pm 0.06}}$ & 75.96$_{{\pm 0.19}}$ & 75.95$_{{\pm 0.19}}$ & 68.21$_{{\pm 0.29}}$ & 61.57$_{{\pm 0.15}}$ \\
 & Long-CoT & 88.30$_{{\pm 0.10}}$ & 88.16$_{{\pm 0.11}}$ & 89.40$_{{\pm 0.20}}$ & 86.84$_{{\pm 0.04}}$ & \textcolor{blue}{\textbf{95.41}}$_{{\pm 0.00}}$ & \textcolor{blue}{\textbf{95.41}}$_{{\pm 0.00}}$ & 75.45$_{{\pm 0.19}}$ & 75.45$_{{\pm 0.19}}$ & 67.32$_{{\pm 0.18}}$ & 61.06$_{{\pm 0.12}}$ \\
\bottomrule
\end{tabular}%
}
\end{table*}
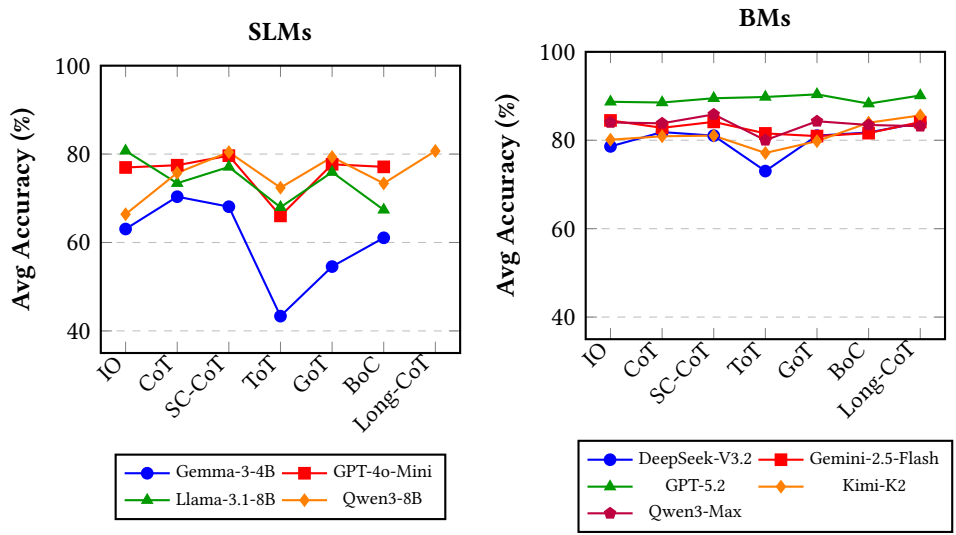
\begin{figure*}[t]
\centering
\begin{minipage}{0.4\textwidth}
  \begin{tikzpicture}
    \begin{axis}[
        title={\textbf{SLMs}},
        width=\linewidth,
        height=0.85\linewidth,
        xlabel={\textbf{Reasoning Method}},
        ylabel={\textbf{Avg Accuracy (\%)}},
        enlarge x limits=0.08,
        symbolic x coords={IO,CoT,SC-CoT,ToT,GoT,BoC,Long-CoT},
        xtick={IO,CoT,SC-CoT,ToT,GoT,BoC,Long-CoT},
        x tick label style={rotate=45, anchor=east, font=\small},
        ymin=35, ymax=100,
        legend style={at={(0.5,-0.35)}, anchor=north, legend columns=2, font=\scriptsize, cells={align=left}},
        ymajorgrids=true,
        grid style=dashed,
        thick,
        mark size=2pt
    ]
        \addplot[color=blue, mark=*]
            coordinates { (IO, 63.05) (CoT, 70.36) (SC-CoT, 68.10) (ToT, 43.34) (GoT, 54.55) (BoC, 61.06) };
        \addlegendentry{Gemma-3-4B}
        \addplot[color=red, mark=square*]
            coordinates { (IO, 76.97) (CoT, 77.48) (SC-CoT, 79.64) (ToT, 66.02) (GoT, 77.69) (BoC, 77.11) };
        \addlegendentry{GPT-4o-Mini}
        \addplot[color=green!60!black, mark=triangle*]
            coordinates { (IO, 80.73) (CoT, 73.41) (SC-CoT, 77.09) (ToT, 67.97) (GoT, 75.91) (BoC, 67.38) };
        \addlegendentry{Llama-3.1-8B}
        \addplot[color=orange, mark=diamond*]
            coordinates { (IO, 66.39) (CoT, 75.75) (SC-CoT, 80.44) (ToT, 72.40) (GoT, 79.33) (BoC, 73.37) (Long-CoT, 80.70) };
        \addlegendentry{Qwen3-8B}
    \end{axis}
  \end{tikzpicture}
\end{minipage}
\begin{minipage}{0.4\textwidth}
  \begin{tikzpicture}
    \begin{axis}[
        title={\textbf{BMs}},
        width=\linewidth,
        height=0.85\linewidth,
        xlabel={\textbf{Reasoning Method}},
        ylabel={\textbf{Avg Accuracy (\%)}},
        enlarge x limits=0.08,
        symbolic x coords={IO,CoT,SC-CoT,ToT,GoT,BoC,Long-CoT},
        xtick={IO,CoT,SC-CoT,ToT,GoT,BoC,Long-CoT},
        x tick label style={rotate=45, anchor=east, font=\small},
        ymin=35, ymax=100,
        legend style={at={(0.5,-0.35)}, anchor=north, legend columns=2, font=\scriptsize, cells={align=left}},
        ymajorgrids=true,
        grid style=dashed,
        thick,
        mark size=2pt
    ]
        \addplot[color=blue, mark=*]
            coordinates { (IO, 78.61) (CoT, 81.85) (SC-CoT, 81.04) (ToT, 73.03) (GoT, 81.04) (BoC, 81.76) (Long-CoT, 83.95) };
        \addlegendentry{DeepSeek-V3.2}
        \addplot[color=red, mark=square*]
            coordinates { (IO, 84.48) (CoT, 82.81) (SC-CoT, 84.16) (ToT, 81.55) (GoT, 80.94) (BoC, 81.61) (Long-CoT, 84.11) };
        \addlegendentry{Gemini-2.5-Flash}
        \addplot[color=green!60!black, mark=triangle*]
            coordinates { (IO, 88.70) (CoT, 88.54) (SC-CoT, 89.50) (ToT, 89.79) (GoT, 90.38) (BoC, 88.29) (Long-CoT, 90.13) };
        \addlegendentry{GPT-5.2}
        \addplot[color=orange, mark=diamond*]
            coordinates { (IO, 80.12) (CoT, 80.90) (SC-CoT, 81.05) (ToT, 77.13) (GoT, 79.80) (BoC, 83.97) (Long-CoT, 85.62) };
        \addlegendentry{Kimi-K2}
        \addplot[color=purple, mark=pentagon*]
            coordinates { (IO, 84.03) (CoT, 83.83) (SC-CoT, 85.85) (ToT, 80.01) (GoT, 84.30) (BoC, 83.46) (Long-CoT, 83.18) };
        \addlegendentry{Qwen3-Max}
    \end{axis}
  \end{tikzpicture}
\end{minipage}
\caption{Average accuracy results across reasoning methods for SLMs (left) and BMs (right).}
\label{fig:method_trend_split}
\end{figure*}
\subsection{Main Results on Big Models}
Tab.~\ref{tab:large_models} presents the performance of different reasoning strategies on big models. 
We focus our analysis on Macro-F1 (Ma-F1), which provides a more robust evaluation under class imbalance. 
Compared with small models, big models exhibit more stable and structured gains from reasoning, while the effectiveness of reasoning remains strongly conditioned on task type, reasoning complexity, and model characteristics.

\textbf{(1) Big models can more reliably convert reasoning into performance gains.}
A key difference from small models is that big models consistently benefit from reasoning, particularly from deeper reasoning strategies. 
For example, DeepSeek-V3.2 improves from 81.45 (IO) to 86.32 (Long-CoT) on AG News (+5.98\%), from 71.47 to 78.19 on SemEval-2018 (+9.40\%), and from 53.68 to 63.95 on iSarcasmEval (+19.13\%). 
Similarly, Kimi-K2 improves from 71.61 to 80.06 on SemEval-2018 (+11.80\%) and from 56.41 to 68.93 on iSarcasmEval (+22.20\%) under Long-CoT. 
These results indicate that stronger models are better able to sustain long reasoning chains and transform additional reasoning budget into effective decision signals.

\textbf{(2) Reasoning is not universally beneficial.}
Despite the overall gains, reasoning does not consistently outperform direct prediction. 
For GPT-5, IO achieves the best performance on AG News (87.86 Ma-F1), outperforming all reasoning variants. 
For Gemini-2.5-Flash, IO is also optimal on AG News (85.31) and SemEval-2018 (79.33). 
Similarly, Kimi-K2 achieves its best result on SST-2 with IO (94.72), while most reasoning strategies lead to slight degradation. 
These findings suggest that for tasks with clear decision boundaries, explicit reasoning may introduce unnecessary intermediate steps without improving classification outcomes.

\textbf{(3) On objective tasks, lightweight reasoning is more robust, while complex reasoning remains risky.}
On objective classification datasets such as AG News and TREC-QC, simple reasoning strategies tend to be more stable and effective. 
In particular, BoC achieves the best TREC-QC performance for multiple models, including DeepSeek-V3.2 (87.84), Gemini-2.5-Flash (87.84), Kimi-K2 (90.22), and Qwen3-Max (89.95). 
Compared to IO, these gains are moderate but consistent (e.g., Kimi-K2: 88.16 $\rightarrow$ 90.22, +2.34\%; DeepSeek-V3.2: 83.59 $\rightarrow$ 87.84, +5.08\%). 

In contrast, ToT remains highly unstable even for big models. 
For DeepSeek-V3.2, ToT reduces AG News from 81.45 to 65.35 (-19.77\%), TREC-QC from 83.59 to 68.37 (-18.21\%), and SST-2 from 94.02 to 83.68 (-11.00\%). 
Similar degradations are observed for Qwen3-Max (e.g., AG News: -12.30\%, TREC-QC: -10.56\%). 
This suggests that structured search introduces excessive branching noise and error propagation when the task does not inherently require multi-step exploration.

\textbf{(4) Reasoning is more effective for subjective tasks, but the optimal strategy is model-dependent.}
For subjective language understanding tasks such as SemEval-2018 and iSarcasmEval, reasoning provides more substantial gains. 
However, unlike small models, no single reasoning strategy dominates across all big models. 
For example, GPT-5 achieves its best SemEval-2018 performance with GoT (88.78), while Long-CoT is optimal on iSarcasmEval (79.35). 
DeepSeek-V3.2 and Kimi-K2 both benefit most from Long-CoT on these tasks, whereas Qwen3-Max achieves its best results with SC-CoT (80.51 on SemEval-2018 and 67.43 on iSarcasmEval). 
These results indicate that reasoning helps capture implicit semantics and pragmatic cues, but the optimal reasoning topology depends on how each model internally organizes contextual inference.

\textbf{(5) Long-CoT becomes effective only at large-model scale, but its benefits remain selective.}
A major contrast with small models is that Long-CoT becomes substantially more useful at large scale. 
It achieves the best performance for DeepSeek-V3.2 on AG News, SemEval-2018, and iSarcasmEval, for GPT-5 on TREC-QC and iSarcasmEval, and for Kimi-K2 on subjective tasks. 
However, Long-CoT is still not universally optimal. 
For instance, it underperforms IO on AG News for GPT-5 and Gemini-2.5-Flash, and is inferior to BoC on TREC-QC for several models. 
Moreover, it is outperformed by SC-CoT on Qwen3-Max for both SemEval-2018 and iSarcasmEval. 
Therefore, the effectiveness of longer reasoning is conditional: it emerges only when the model capacity is sufficient to utilize extended reasoning depth, and even then remains task- and model-dependent.

\textbf{Summary.}
Overall, the results on large models reveal a more nuanced picture of reasoning in text classification. 
While stronger models can better exploit reasoning—especially for subjective tasks—reasoning does not provide universal gains. 
Direct prediction remains competitive on many tasks, and overly complex reasoning (e.g., ToT) can still degrade performance. 
These findings suggest that reasoning is not a monotonic function of performance, but a conditional resource whose utility depends on the alignment among task complexity, reasoning structure, and model capacity.

\subsection{Comparison Between Small and Big Models}
Fig.~\ref{fig:method_trend_split} compares the average accuracy of different reasoning strategies across datasets for small and large models. 
A clear scaling pattern emerges: reasoning becomes more stable and consistently beneficial as model size increases. 
For small models, reasoning gains are often unstable and can even be detrimental. 
For example, ToT reduces Gemma-3-4B from 63.05 (IO) to 43.34 (-31.25\%), and performance varies widely across methods. 
In contrast, large models exhibit more consistent improvements: DeepSeek-V3.2 improves from 78.61 (IO) to 83.95 (Long-CoT, +6.79\%), and Kimi-K2 from 80.12 to 85.62 (+6.86\%). 
Meanwhile, performance variance across reasoning methods shrinks significantly (e.g., GPT-5.2 ranges only from 88.29 to 90.38), indicating that larger models are more robust to different reasoning paradigms.

Another key difference lies in the effectiveness of reasoning complexity. 
Long-CoT provides limited gains for small models (e.g., Qwen3-8B: 80.44 to 80.70, +0.32\%), but becomes one of the strongest strategies for large models, achieving the best or near-best results across multiple systems (e.g., DeepSeek-V3.2: 83.95; Kimi-K2: 85.62). 
In contrast, ToT remains consistently unstable across both scales, although its negative impact is less severe for large models (e.g., -31.25\% vs.\ -7.11\%). 
These results suggest that reasoning effectiveness does not grow monotonically with complexity, but depends on the alignment between reasoning depth, structure, and model capacity.

\begin{table*}[t]
\centering
\caption{Efficiency of different reasoning methods on \textbf{SLMs}. \textcolor{red}{\textbf{Red}} indicates the overall best \textit{AE} in each column, while \textcolor{blue}{\textbf{blue}} indicates the best \textit{AE} for each individual model. \textit{AE} is scaled by $10^6$ and \textit{ME} is scaled by $10^8$. Superscripts indicate the efficiency regime (see Table \ref{tab:me_interpretation}).}
\label{tab:efficiency_small_models}
\resizebox{\textwidth}{!}{%
\begin{tabular}{llcccccccccc}
\toprule
\multirow{2}{*}{\textbf{Model}} & \multirow{2}{*}{\textbf{Method}} & \multicolumn{2}{c}{\textbf{AG News}} & \multicolumn{2}{c}{\textbf{TREC-QC}} & \multicolumn{2}{c}{\textbf{SST-2}} & \multicolumn{2}{c}{\textbf{SemEval-2018}} & \multicolumn{2}{c}{\textbf{iSarcasmEval}} \\
\cmidrule(lr){3-4} \cmidrule(lr){5-6} \cmidrule(lr){7-8} \cmidrule(lr){9-10} \cmidrule(lr){11-12}
 &  & \makebox[1.3cm][c]{AE} & \makebox[1.5cm][c]{ME} & \makebox[1.3cm][c]{AE} & \makebox[1.5cm][c]{ME} & \makebox[1.3cm][c]{AE} & \makebox[1.5cm][c]{ME} & \makebox[1.3cm][c]{AE} & \makebox[1.5cm][c]{ME} & \makebox[1.3cm][c]{AE} & \makebox[1.5cm][c]{ME} \\
\midrule
\multirow{6}{*}{GPT-4o-mini} & IO & \textcolor{blue}{\textbf{10.66}} & - & \textcolor{blue}{\textbf{5.50}} & - & \textcolor{blue}{\textbf{4.55}} & - & \textcolor{blue}{\textbf{3.40}} & - & \textcolor{blue}{\textbf{1.38}} & - \\
 & CoT & 5.08 & 42.70$^{\bigstar}$ & 3.09 & -11.41$^{\spadesuit}$ & 2.56 & 2.91$^{\bigstar}$ & 1.75 & -4.15$^{\spadesuit}$ & 0.92 & 1.76$^{\bigstar}$ \\
 & SC-CoT & 0.79 & 3.75$^{\bigstar}$ & 0.95 & -3.48$^{\spadesuit}$ & 0.68 & 0.83$^{\bigstar}$ & 0.51 & 3.79$^{\bigstar}$ & 0.26 & 3.16$^{\bigstar}$ \\
 & ToT & 0.32 & -24.16$^{\spadesuit}$ & 0.36 & -24.29$^{\spadesuit}$ & 0.37 & -5.34$^{\spadesuit}$ & 0.35 & 4.54$^{\bigstar}$ & 0.16 & 2.74$^{\bigstar}$ \\
 & GoT & 0.57 & -0.50$^{\spadesuit}$ & 0.60 & -6.99$^{\spadesuit}$ & 0.45 & -1.15$^{\spadesuit}$ & 0.36 & 2.58$^{\bigstar}$ & 0.16 & 1.69$^{\bigstar}$ \\
 & BoC & 2.30 & 2.39$^{\bigstar}$ & 1.39 & -5.85$^{\spadesuit}$ & 1.43 & 1.24$^{\bigstar}$ & 1.14 & 3.08$^{\bigstar}$ & 0.62 & 1.26$^{\bigstar}$ \\
\midrule
\multirow{6}{*}{Gemma-3-4B-IT} & IO & \textcolor{red}{\textbf{88.34}} & - & \textcolor{blue}{\textbf{37.70}} & - & \textcolor{blue}{\textbf{29.35}} & - & \textcolor{blue}{\textbf{11.77}} & - & \textcolor{blue}{\textbf{4.76}} & - \\
 & CoT & 5.15 & 25.51$^{\bigstar}$ & 1.77 & -17.99$^{\spadesuit}$ & 1.48 & 0.18$^{\bigstar}$ & 0.82 & 28.07$^{\bigstar}$ & 0.36 & 14.90$^{\bigstar}$ \\
 & SC-CoT & 1.40 & 9.48$^{\bigstar}$ & 0.56 & 2.51$^{\bigstar}$ & 0.31 & -0.22$^{\spadesuit}$ & 0.18 & 4.45$^{\bigstar}$ & 0.06 & 1.68$^{\bigstar}$ \\
 & ToT & 0.12 & -10.99$^{\spadesuit}$ & 0.04 & -4.34$^{\spadesuit}$ & 0.05 & -0.52$^{\spadesuit}$ & 0.02 & -0.34$^{\spadesuit}$ & 0.01 & -0.28$^{\spadesuit}$ \\
 & GoT & 0.26 & -2.47$^{\spadesuit}$ & 0.13 & -1.23$^{\spadesuit}$ & 0.06 & -1.56$^{\spadesuit}$ & 0.03 & -0.58$^{\spadesuit}$ & 0.01 & -0.38$^{\spadesuit}$ \\
 & BoC & 17.67 & 21.97$^{\bigstar}$ & 6.98 & -31.79$^{\spadesuit}$ & 2.66 & -0.46$^{\spadesuit}$ & 1.49 & -6.49$^{\spadesuit}$ & 0.24 & -2.97$^{\spadesuit}$ \\
\midrule
\multirow{6}{*}{Llama-3.1-8B-Instruct} & IO & \textcolor{blue}{\textbf{80.37}} & - & \textcolor{blue}{\textbf{51.79}} & - & \textcolor{red}{\textbf{33.42}} & - & \textcolor{red}{\textbf{18.56}} & - & \textcolor{red}{\textbf{9.14}} & - \\
 & CoT & 3.78 & 3.16$^{\bigstar}$ & 2.08 & -34.62$^{\spadesuit}$ & 1.55 & -0.69$^{\spadesuit}$ & 0.96 & -13.49$^{\spadesuit}$ & 0.47 & -9.52$^{\spadesuit}$ \\
 & SC-CoT & 0.45 & -0.17$^{\spadesuit}$ & 0.29 & 0.01$^{\bigstar}$ & 0.24 & 0.35$^{\bigstar}$ & 0.17 & -1.07$^{\spadesuit}$ & 0.08 & -1.07$^{\spadesuit}$ \\
 & ToT & 0.21 & -2.36$^{\spadesuit}$ & 0.16 & -2.86$^{\spadesuit}$ & 0.11 & -0.95$^{\spadesuit}$ & 0.08 & -1.12$^{\spadesuit}$ & 0.04 & -0.92$^{\spadesuit}$ \\
 & GoT & 0.25 & 0.18$^{\bigstar}$ & 0.21 & -1.60$^{\spadesuit}$ & 0.13 & 0.04$^{\bigstar}$ & 0.11 & -0.63$^{\spadesuit}$ & 0.05 & -0.66$^{\spadesuit}$ \\
 & BoC & 10.57 & -63.44$^{\spadesuit}$ & 9.85 & -11.85$^{\spadesuit}$ & 6.34 & 0.66$^{\bigstar}$ & 3.63 & -231.27$^{\spadesuit}$ & 1.45 & -178.78$^{\spadesuit}$ \\
\midrule
\multirow{7}{*}{Qwen3-8B} & IO & \textcolor{blue}{\textbf{78.39}} & - & \textcolor{red}{\textbf{59.93}} & - & 5.79 & - & 3.25 & - & \textcolor{blue}{\textbf{8.03}} & - \\
 & CoT & 18.06 & 154.53$^{\bigstar}$ & 15.69 & -168.72$^{\spadesuit}$ & 5.09 & -133.07$^{\spadesuit}$ & 3.40 & 374.97$^{\bigstar}$ & 3.49 & 187.74$^{\bigstar}$ \\
 & SC-CoT & 2.34 & 10.95$^{\bigstar}$ & 3.11 & -8.74$^{\spadesuit}$ & 1.49 & -2.22$^{\spadesuit}$ & 1.09 & 56.08$^{\bigstar}$ & 0.62 & 30.62$^{\bigstar}$ \\
 & ToT & 0.47 & -10.84$^{\spadesuit}$ & 0.43 & -22.57$^{\spadesuit}$ & 0.37 & -4.89$^{\spadesuit}$ & 0.35 & 16.25$^{\bigstar}$ & 0.18 & 9.38$^{\bigstar}$ \\
 & GoT & 1.51 & 10.35$^{\bigstar}$ & 1.69 & -3.81$^{\spadesuit}$ & 0.98 & -2.08$^{\spadesuit}$ & 0.75 & 34.31$^{\bigstar}$ & 0.38 & 17.72$^{\bigstar}$ \\
 & BoC & 14.12 & -45.48$^{\spadesuit}$ & 17.13 & -134.88$^{\spadesuit}$ & \textcolor{blue}{\textbf{8.54}} & -17.17$^{\clubsuit}$ & \textcolor{blue}{\textbf{7.79}} & -328.85$^{\clubsuit}$ & 2.39 & 107.36$^{\bigstar}$ \\
 & Long-CoT & 6.27 & 32.76$^{\bigstar}$ & 5.60 & 12.83$^{\bigstar}$ & 3.85 & -5.15$^{\spadesuit}$ & 2.35 & 165.46$^{\bigstar}$ & 1.08 & 54.48$^{\bigstar}$ \\
\bottomrule
\end{tabular}%
}
\end{table*}

\subsection{Efficiency Analysis on Small Models}
We analyze efficiency from two complementary perspectives: AE, which measures overall cost-effectiveness, and ME, which captures the incremental benefit of reasoning relative to the IO baseline. 
Overall, a consistent pattern emerges: while reasoning occasionally provides positive marginal gains, it rarely improves overall efficiency on small models.

First, from the perspective of AE, the IO baseline dominates almost all settings. 
Across 4 models and 5 datasets (20 configurations), IO achieves the highest AE in 18 cases. 
The only exceptions occur for Qwen3-8B on SST-2 and SemEval-2018, where BoC surpasses IO (8.54 vs.\ 5.79 and 7.79 vs.\ 3.25, respectively). 
In all other cases, reasoning strategies lead to substantially lower AE. 
For example, on Gemma-3-4B-IT, IO achieves an AE of 88.34 on AG News, whereas CoT and SC-CoT drop to 5.15 and 1.40. 
Similarly, for Llama-3.1-8B-Instruct, IO reaches 18.56 and 9.14 on SemEval-2018 and iSarcasmEval, while all reasoning methods remain significantly lower. 
These results indicate that the additional token cost introduced by reasoning is rarely compensated by proportional performance gains, making IO the most cost-effective strategy overall.

However, ME reveals a more nuanced picture. 
Although reasoning is inefficient in absolute terms, it can still yield positive marginal returns in specific cases. 
For instance, GPT-4o-mini achieves a ME of 42.70 with CoT on AG News, and Gemma-3-4B-IT obtains ME values of 28.07 and 14.90 with CoT on SemEval-2018 and iSarcasmEval, respectively. 
This effect is particularly pronounced for Qwen3-8B, where CoT reaches extremely high ME values on SemEval-2018 (374.97) and iSarcasmEval (187.74), and Long-CoT also achieves 165.46 and 54.48 on these datasets. 
Nevertheless, these methods still exhibit much lower AE than IO, indicating that while additional reasoning may be worthwhile in a marginal sense, it does not translate into superior overall efficiency. 
This highlights a key distinction: ME reflects whether extra reasoning is beneficial, whereas AE determines whether it is ultimately cost-effective.

From a task perspective, positive ME values are concentrated in subjective language understanding tasks, while objective classification tasks are dominated by negative ME. 
On TREC-QC and SST-2, most reasoning methods yield negative ME (e.g., for GPT-4o-mini on TREC-QC: -11.41, -3.48, -24.29, -6.99, -5.85 across methods), indicating that additional reasoning increases cost without sufficient performance improvement. 
In contrast, on SemEval-2018 and iSarcasmEval, positive ME is much more frequent, particularly for GPT-4o-mini, Gemma-3-4B-IT, and Qwen3-8B. 
A notable case is Qwen3-8B with BoC, which achieves \(\Delta \text{F1} > 0\) and \(\Delta T < 0\) on SST-2 and SemEval-2018, corresponding to a Pareto improvement. 
Although ME is negative in this regime, the method simultaneously improves performance and reduces cost, demonstrating that ME must be interpreted jointly with the underlying efficiency regime.

Finally, from a model perspective, small models exhibit highly heterogeneous responses to reasoning, but share a common limitation: complex reasoning strategies are consistently inefficient. 
For GPT-4o-mini, ToT achieves extremely low AE (e.g., 0.32, 0.36, 0.37 across datasets) and strongly negative ME, indicating severe inefficiency. 
Gemma-3-4B-IT is even more sensitive, with ToT and GoT producing AE values close to zero across all tasks. 
Llama-3.1-8B-Instruct shows the strongest resistance to reasoning, with IO dominating all settings and BoC yielding extremely negative ME (e.g., -231.27 on SemEval-2018 and -178.78 on iSarcasmEval). 

Overall, the joint analysis of AE and ME reveals that, for small models, reasoning is not a generally cost-effective strategy.

\begin{table*}[t]
\centering
\caption{Efficiency of different reasoning methods on \textbf{BMs}. \textcolor{red}{\textbf{Red}} indicates the overall best AE in each column, while \textcolor{blue}{\textbf{blue}} indicates the best AE for each individual model. AE is scaled by $10^6$ and ME is scaled by $10^8$. Superscripts indicate the efficiency regime (see Table \ref{tab:me_interpretation}).}
\label{{tab:efficiency_large_models}}
\resizebox{\textwidth}{!}{%
\begin{tabular}{llcccccccccc}
\toprule
\multirow{2}{*}{\textbf{Model}} & \multirow{2}{*}{\textbf{Method}} & \multicolumn{2}{c}{\textbf{AG News (subset)}} & \multicolumn{2}{c}{\textbf{TREC-QC}} & \multicolumn{2}{c}{\textbf{SST-2}} & \multicolumn{2}{c}{\textbf{SemEval-2018}} & \multicolumn{2}{c}{\textbf{iSarcasmEval}} \\
\cmidrule(lr){3-4} \cmidrule(lr){5-6} \cmidrule(lr){7-8} \cmidrule(lr){9-10} \cmidrule(lr){11-12}
 &  & \makebox[1.3cm][c]{AE} & \makebox[1.5cm][c]{ME} & \makebox[1.3cm][c]{AE} & \makebox[1.5cm][c]{ME} & \makebox[1.3cm][c]{AE} & \makebox[1.5cm][c]{ME} & \makebox[1.3cm][c]{AE} & \makebox[1.5cm][c]{ME} & \makebox[1.3cm][c]{AE} & \makebox[1.5cm][c]{ME} \\
\midrule
\multirow{7}{*}{DeepSeek-V3.2} & IO & \textcolor{blue}{\textbf{32.49}} & - & \textcolor{blue}{\textbf{28.92}} & - & \textcolor{blue}{\textbf{18.27}} & - & \textcolor{blue}{\textbf{15.72}} & - & \textcolor{blue}{\textbf{6.05}} & - \\
 & CoT & 6.20 & 26.21$^{\bigstar}$ & 7.61 & 38.24$^{\bigstar}$ & 4.69 & 0.71$^{\bigstar}$ & 3.15 & 25.19$^{\bigstar}$ & 1.44 & 20.90$^{\bigstar}$ \\
 & SC-CoT & 1.36 & 5.97$^{\bigstar}$ & 1.34 & 6.76$^{\bigstar}$ & 0.87 & 1.28$^{\bigstar}$ & 0.52 & 0.29$^{\bigstar}$ & 0.22 & 1.92$^{\bigstar}$ \\
 & ToT & 0.33 & -8.23$^{\spadesuit}$ & 0.33 & -7.41$^{\spadesuit}$ & 0.25 & -3.17$^{\spadesuit}$ & 0.21 & 0.79$^{\bigstar}$ & 0.10 & 1.18$^{\bigstar}$ \\
 & GoT & 0.63 & -0.99$^{\spadesuit}$ & 0.60 & 1.22$^{\bigstar}$ & 0.33 & -0.26$^{\spadesuit}$ & 0.26 & 1.36$^{\bigstar}$ & 0.12 & 1.73$^{\bigstar}$ \\
 & BoC & 2.85 & 9.25$^{\bigstar}$ & 6.26 & 38.16$^{\bigstar}$ & 1.56 & 1.27$^{\bigstar}$ & 1.40 & 7.89$^{\bigstar}$ & 0.56 & 6.19$^{\bigstar}$ \\
 & Long-CoT & 2.85 & 17.50$^{\bigstar}$ & 2.35 & 7.22$^{\bigstar}$ & 2.36 & 2.50$^{\bigstar}$ & 1.10 & 10.13$^{\bigstar}$ & 0.47 & 8.02$^{\bigstar}$ \\
\midrule
\multirow{7}{*}{GPT-5} & IO & \textcolor{blue}{\textbf{22.37}} & - & \textcolor{blue}{\textbf{17.64}} & - & \textcolor{blue}{\textbf{8.58}} & - & \textcolor{blue}{\textbf{11.74}} & - & \textcolor{blue}{\textbf{5.66}} & - \\
 & CoT & 8.63 & -19.11$^{\spadesuit}$ & 11.27 & -10.65$^{\spadesuit}$ & 4.41 & 0.47$^{\bigstar}$ & 5.35 & 1.33$^{\bigstar}$ & 2.58 & 1.92$^{\bigstar}$ \\
 & SC-CoT & 1.69 & -0.11$^{\spadesuit}$ & 1.93 & 0.59$^{\bigstar}$ & 0.88 & 0.27$^{\bigstar}$ & 0.79 & 0.64$^{\bigstar}$ & 0.36 & 0.47$^{\bigstar}$ \\
 & ToT & 1.10 & -0.86$^{\spadesuit}$ & 1.03 & 0.17$^{\bigstar}$ & 0.65 & 0.49$^{\bigstar}$ & 0.65 & 1.46$^{\bigstar}$ & 0.30 & 0.55$^{\bigstar}$ \\
 & GoT & 1.03 & -0.88$^{\spadesuit}$ & 1.06 & -0.09$^{\spadesuit}$ & 0.66 & 0.29$^{\bigstar}$ & 0.63 & 2.17$^{\bigstar}$ & 0.29 & 1.66$^{\bigstar}$ \\
 & BoC & 4.33 & -3.05$^{\spadesuit}$ & 4.97 & -8.92$^{\spadesuit}$ & 2.58 & 1.13$^{\bigstar}$ & 2.42 & -1.86$^{\spadesuit}$ & 1.12 & -4.05$^{\spadesuit}$ \\
 & Long-CoT & 6.65 & -1.61$^{\spadesuit}$ & 5.13 & 11.40$^{\bigstar}$ & 3.71 & 3.88$^{\bigstar}$ & 2.87 & 8.31$^{\bigstar}$ & 1.29 & 9.15$^{\bigstar}$ \\
\midrule
\multirow{7}{*}{Gemini-2.5-Flash} & IO & \textcolor{blue}{\textbf{4.46}} & - & \textcolor{blue}{\textbf{4.60}} & - & \textcolor{blue}{\textbf{2.89}} & - & \textcolor{blue}{\textbf{2.04}} & - & \textcolor{blue}{\textbf{0.86}} & - \\
 & CoT & 2.72 & -5.26$^{\spadesuit}$ & 3.29 & 4.06$^{\bigstar}$ & 2.07 & -0.59$^{\spadesuit}$ & 1.37 & -29.54$^{\spadesuit}$ & 0.62 & -7.61$^{\spadesuit}$ \\
 & SC-CoT & 0.47 & -0.55$^{\spadesuit}$ & 0.68 & -1.88$^{\spadesuit}$ & 0.38 & -0.18$^{\spadesuit}$ & 0.27 & -0.30$^{\spadesuit}$ & 0.12 & 0.06$^{\bigstar}$ \\
 & ToT & 0.26 & -0.16$^{\spadesuit}$ & 0.27 & -4.35$^{\spadesuit}$ & 0.20 & -0.37$^{\spadesuit}$ & 0.16 & -0.77$^{\spadesuit}$ & 0.07 & -0.30$^{\spadesuit}$ \\
 & GoT & 0.34 & -1.59$^{\spadesuit}$ & 0.43 & -1.14$^{\spadesuit}$ & 0.24 & -0.21$^{\spadesuit}$ & 0.18 & -1.70$^{\spadesuit}$ & 0.08 & -0.98$^{\spadesuit}$ \\
 & BoC & 0.67 & -0.57$^{\spadesuit}$ & 1.10 & 3.27$^{\bigstar}$ & 0.54 & 0.45$^{\bigstar}$ & 0.34 & -4.00$^{\spadesuit}$ & 0.16 & -2.41$^{\spadesuit}$ \\
 & Long-CoT & 2.44 & -2.36$^{\spadesuit}$ & 2.84 & -11.78$^{\spadesuit}$ & 1.82 & 1.00$^{\bigstar}$ & 0.92 & -3.65$^{\spadesuit}$ & 0.42 & -0.85$^{\spadesuit}$ \\
\midrule
\multirow{7}{*}{Kimi-K2} & IO & \textcolor{red}{\textbf{127.52}} & - & \textcolor{red}{\textbf{123.48}} & - & \textcolor{red}{\textbf{104.65}} & - & \textcolor{red}{\textbf{76.80}} & - & \textcolor{red}{\textbf{31.46}} & - \\
 & CoT & 9.15 & 23.80$^{\bigstar}$ & 10.93 & 6.67$^{\bigstar}$ & 7.01 & -8.10$^{\spadesuit}$ & 4.11 & -5.02$^{\spadesuit}$ & 2.00 & 2.79$^{\bigstar}$ \\
 & SC-CoT & 1.87 & 5.10$^{\bigstar}$ & 2.06 & 4.42$^{\bigstar}$ & 1.21 & -0.30$^{\spadesuit}$ & 0.78 & -0.84$^{\spadesuit}$ & 0.31 & 0.22$^{\bigstar}$ \\
 & ToT & 0.81 & -5.11$^{\spadesuit}$ & 0.66 & -12.21$^{\spadesuit}$ & 0.47 & -3.56$^{\spadesuit}$ & 0.37 & 0.85$^{\bigstar}$ & 0.17 & 1.02$^{\bigstar}$ \\
 & GoT & 0.77 & -2.72$^{\spadesuit}$ & 0.86 & -3.85$^{\spadesuit}$ & 0.48 & -1.14$^{\spadesuit}$ & 0.36 & 0.75$^{\bigstar}$ & 0.16 & 0.97$^{\bigstar}$ \\
 & BoC & 4.90 & 12.14$^{\bigstar}$ & 5.22 & 12.41$^{\bigstar}$ & 2.82 & -1.58$^{\spadesuit}$ & 2.47 & 13.14$^{\bigstar}$ & 1.30 & 17.81$^{\bigstar}$ \\
 & Long-CoT & 2.93 & 7.36$^{\bigstar}$ & 3.11 & -1.93$^{\spadesuit}$ & 2.15 & -1.88$^{\spadesuit}$ & 1.45 & 15.54$^{\bigstar}$ & 0.61 & 11.20$^{\bigstar}$ \\
\midrule
\multirow{7}{*}{Qwen3-Max} & IO & \textcolor{blue}{\textbf{37.92}} & - & \textcolor{blue}{\textbf{41.83}} & - & \textcolor{blue}{\textbf{25.69}} & - & \textcolor{blue}{\textbf{22.74}} & - & \textcolor{blue}{\textbf{9.79}} & - \\
 & CoT & 11.27 & 13.10$^{\bigstar}$ & 10.07 & 16.13$^{\bigstar}$ & 6.84 & -1.36$^{\spadesuit}$ & 4.39 & -13.62$^{\spadesuit}$ & 2.01 & 2.54$^{\bigstar}$ \\
 & SC-CoT & 1.55 & 1.37$^{\bigstar}$ & 1.38 & 1.84$^{\bigstar}$ & 0.92 & 0.50$^{\bigstar}$ & 0.63 & 1.23$^{\bigstar}$ & 0.32 & 2.71$^{\bigstar}$ \\
 & ToT & 0.44 & -6.22$^{\spadesuit}$ & 0.40 & -4.72$^{\spadesuit}$ & 0.29 & -1.34$^{\spadesuit}$ & 0.22 & -1.74$^{\spadesuit}$ & 0.11 & 0.62$^{\bigstar}$ \\
 & GoT & 1.17 & -0.13$^{\spadesuit}$ & 1.11 & 1.40$^{\bigstar}$ & 0.63 & -0.57$^{\spadesuit}$ & 0.51 & 0.01$^{\bigstar}$ & 0.24 & 0.76$^{\bigstar}$ \\
 & BoC & 2.89 & -0.83$^{\spadesuit}$ & 7.58 & 15.50$^{\bigstar}$ & 1.51 & -0.73$^{\spadesuit}$ & 2.07 & -9.14$^{\spadesuit}$ & 0.97 & -0.51$^{\spadesuit}$ \\
 & Long-CoT & 21.96 & 16.34$^{\bigstar}$ & 29.73 & -196.99$^{\spadesuit}$ & 14.68 & 23.64$^{\bigstar}$ & 7.56 & -54.26$^{\spadesuit}$ & 3.34 & -6.73$^{\spadesuit}$ \\
\bottomrule
\end{tabular}%
}
\end{table*}
\subsection{Efficiency Analysis on Big Models}
We extend the efficiency analysis to big models and observe a fundamentally different pattern compared to small models. 
While IO remains the most cost-effective strategy in terms of AE, reasoning strategies exhibit significantly stronger marginal benefits.

First, IO continues to dominate AE across all settings. 
For every model and dataset, IO achieves the highest AE, often by a large margin. 
For example, Kimi-K2 attains extremely high AE values, reaching 127.52 on AG News, 123.48 on TREC-QC, and 104.65 on SST-2, substantially outperforming all reasoning methods. 
Similarly, DeepSeek-V3.2 and Qwen3-Max consistently show the highest AE under IO across all datasets. 
These results indicate that, even for big models, the additional token cost of reasoning is generally not offset by proportional performance gains, and IO remains the most cost-efficient choice overall.

However, compared to small models, reasoning strategies on big models exhibit substantially improved ME, suggesting stronger returns on additional computation. 
For instance, DeepSeek-V3.2 achieves consistently positive ME across almost all tasks and methods (e.g., CoT: 26.21–38.24; BoC: up to 38.16), indicating stable gains from reasoning. 
Similarly, Kimi-K2 shows strong positive ME for CoT and BoC on multiple datasets (e.g., BoC: 12.14 on AG News, 13.14 on SemEval-2018, and 17.81 on iSarcasmEval). 
Even more strikingly, Qwen3-Max demonstrates large positive ME values for Long-CoT on AG News (16.34) and SST-2 (23.64), suggesting that extended reasoning can yield substantial incremental benefits when model capacity is sufficient. 
In contrast to small models, where ME is frequently negative, big models show predominantly positive ME, especially on subjective tasks.

From a task perspective, reasoning is more beneficial for subjective understanding tasks than for objective classification. 
On SemEval-2018 and iSarcasmEval, most reasoning methods yield positive ME across models (e.g., DeepSeek-V3.2 CoT: 25.19 and 20.90; Kimi-K2 BoC: 13.14 and 17.81), indicating that additional reasoning effectively captures implicit semantics and pragmatic cues. 
In contrast, on TREC-QC and SST-2, negative ME still appears in several cases (e.g., GPT-5 CoT: -10.65 on TREC-QC; Gemini-2.5-Flash CoT: -0.59 on SST-2), suggesting that reasoning remains less beneficial when decision boundaries are relatively explicit. 
Thus, the task-dependent nature of reasoning efficiency persists even as model scale increases.

Finally, from a model scaling perspective, a clear transition emerges: bigger models convert additional reasoning cost into useful performance gains more effectively. 
Compared to small models, where complex reasoning (e.g., ToT, GoT) is almost always inefficient, big models exhibit more stable and positive ME even for these methods. 
Nevertheless, their AE remains significantly lower than IO, indicating that improved marginal gains do not fully compensate for the increased computational cost. 
In other words, reasoning becomes more \emph{useful} with scale, but not necessarily more \emph{economical}.

Overall, the joint analysis of AE and ME suggests a scaling law of reasoning efficiency: as model size increases, the marginal value of reasoning improves substantially, but its overall cost-effectiveness remains constrained. 
This highlights a fundamental trade-off between reasoning depth and computational efficiency, even for state-of-the-art large language models.
\begin{figure}[t]
    \centering
    \includegraphics[width=\textwidth]{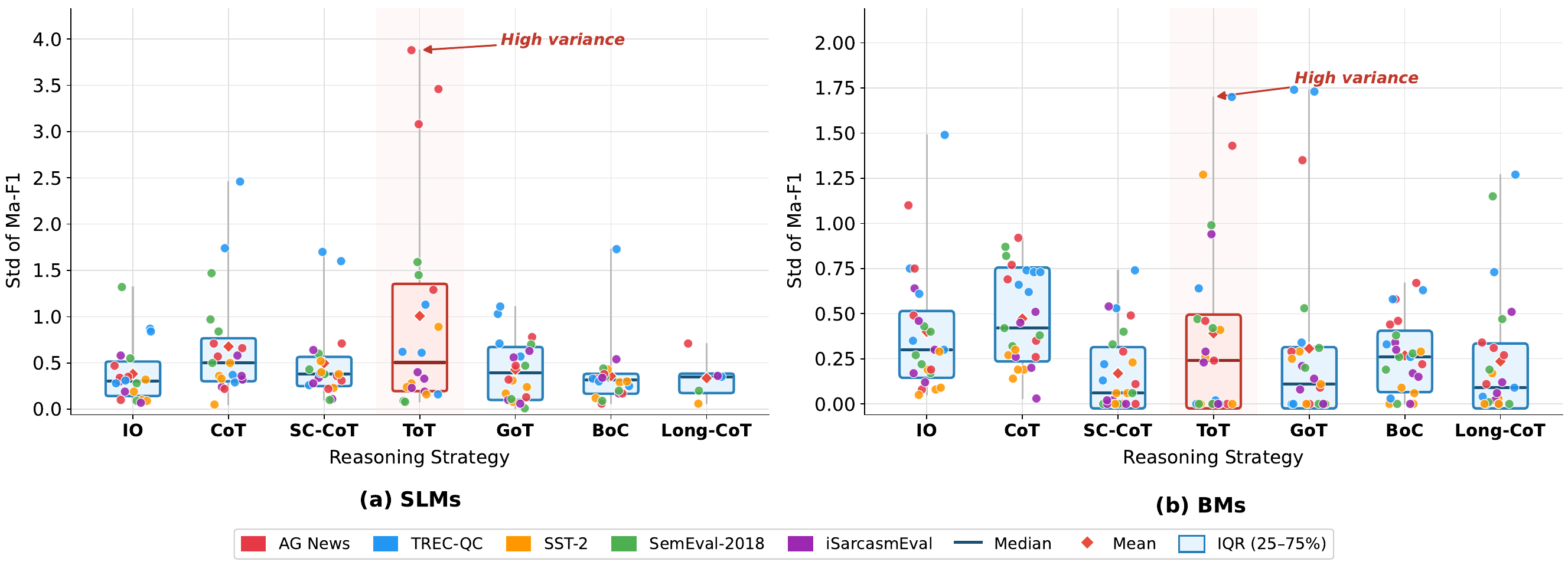}
    \caption{Distribution of Ma-F1 standard deviations across reasoning strategies, models, and datasets for SLMs (a) and BMs (b). Each dot represents one model--dataset combination, colored by dataset.}
    \label{fig:stability}
\end{figure}

\subsection{Stability Analysis}
Figure~\ref{fig:stability} presents the distribution of Ma-F1 standard deviations across all model--dataset combinations. The most prominent pattern is that \textbf{ToT exhibits substantially higher variance than all other strategies} in both SLMs and BMs, with extreme outliers reaching 3.46 and 3.88 in the SLM setting, contributed predominantly by AG News (red dots). The high-variance outliers are disproportionately concentrated on objective multi-class tasks, suggesting that ToT's tree-structured search is particularly prone to path-dependent divergence across runs on these datasets. In contrast, IO and SC-CoT produce the most compact distributions, with data points clustered near zero across nearly all model--dataset combinations, indicating strong reproducibility regardless of model scale or task type.

Comparing the two panels, BMs exhibit markedly lower overall variance than SLMs across all strategies. SC-CoT in particular shows a high concentration of near-zero standard deviations in the BM setting, reflecting the near-deterministic outputs of stronger models under multi-sample voting. Nonetheless, ToT remains the most unstable strategy even among BMs, confirming that its instability is structural rather than capacity-dependent. These results reinforce the practical recommendation that IO and SC-CoT are the most reliable choices for deployment-sensitive classification scenarios.

\subsection{Few-shot Analysis}
Table~\ref{tab:few_shot_comparison} compares zero-shot and 4-shot performance, revealing that few-shot prompting consistently improves performance across most settings, with substantially larger gains observed on smaller models and more complex tasks. 
For GPT-5.2, the improvements are generally modest, typically within 1--3\%. 
For example, CoT improves Macro-F1 on AG News from 86.69 to 88.06 (+1.58\%), and on SST-2 from 95.31 to 96.22 (+0.94\%). 
Larger gains appear on more subjective tasks, such as SemEval-2018 (+3.69\%) and iSarcasmEval (+3.89\%), suggesting that few-shot examples help refine implicit reasoning rather than basic classification boundaries.

In contrast, Qwen3-8B benefits significantly more from few-shot prompting, with improvements often exceeding 10\%. 
For instance, CoT achieves a gain of +12.09 on SemEval-2018 (+18.30\%) and +10.14 on iSarcasmEval (+19.15\%), while IO also shows large improvements, such as +16.63 on SemEval-2018 (+37.79\%) and +10.33 on iSarcasmEval (+32.30\%). 
Notably, few-shot prompting can even outperform reasoning improvements: on AG News, IO improves by +7.75 (+9.90\%), significantly larger than the CoT gain of +1.07 (+1.27\%). 
However, few-shot prompting is not universally beneficial; for example, Qwen3-8B with IO on TREC-QC drops by -4.27\%, indicating potential overfitting to demonstration examples.

Overall, these results suggest that few-shot prompting primarily acts as a strong external prior, particularly for smaller models and subjective tasks, while its marginal benefit diminishes for large models with already well-calibrated decision boundaries.
\begin{table*}[htbp]
\centering
\small
\caption{Zero-Shot vs 4-Shot Performance Comparison (\%)}
\label{tab:few_shot_comparison}
\begin{tabular}{ll|ccc|ccc}
\toprule
\textbf{Dataset} & \textbf{Model (Method)} & \multicolumn{3}{c|}{\textbf{Accuracy}} & \multicolumn{3}{c}{\textbf{Macro-F1}} \\
& & Zero & 4-Shot & $\Delta$ & Zero & 4-Shot & $\Delta$ \\
\midrule
\multirow{4}{*}{AG News (subset)} & GPT-5 {\scriptsize (CoT)} & 86.53 & 88.20 & \textbf{+1.67} & 86.69 & 88.06 & \textbf{+1.37} \\
 & GPT-5 {\scriptsize (IO)} & 87.80 & 88.60 & \textbf{+0.80} & 87.86 & 88.53 & \textbf{+0.67} \\
 & Qwen3-8B {\scriptsize (CoT)} & 84.13 & 85.00 & \textbf{+0.87} & 83.92 & 84.99 & \textbf{+1.07} \\
 & Qwen3-8B {\scriptsize (IO)} & 78.47 & 86.00 & \textcolor{red}{\textbf{+7.53}} & 78.28 & 86.04 & \textcolor{red}{\textbf{+7.75}} \\
\addlinespace
\multirow{4}{*}{TREC-QC} & GPT-5 {\scriptsize (CoT)} & 92.60 & 93.60 & \textbf{+1.00} & 92.26 & 93.38 & \textbf{+1.11} \\
 & GPT-5 {\scriptsize (IO)} & 93.00 & 93.60 & \textbf{+0.60} & 92.58 & 93.38 & \textbf{+0.80} \\
 & Qwen3-8B {\scriptsize (CoT)} & 78.92 & 87.00 & \textcolor{red}{\textbf{+8.08}} & 75.96 & 85.92 & \textcolor{red}{\textbf{+9.96}} \\
 & Qwen3-8B {\scriptsize (IO)} & 80.16 & 77.20 & -2.96 & 81.82 & 78.33 & -3.49 \\
\addlinespace
\multirow{4}{*}{SST-2} & GPT-5 {\scriptsize (CoT)} & 95.18 & 96.22 & \textbf{+1.03} & 95.31 & 96.22 & \textbf{+0.90} \\
 & GPT-5 {\scriptsize (IO)} & 94.53 & 95.99 & \textbf{+1.45} & 95.26 & 96.04 & \textbf{+0.78} \\
 & Qwen3-8B {\scriptsize (CoT)} & 90.64 & 92.78 & \textcolor{red}{\textbf{+2.13}} & 90.62 & 92.77 & \textcolor{red}{\textbf{+2.15}} \\
 & Qwen3-8B {\scriptsize (IO)} & 92.11 & 93.81 & \textbf{+1.70} & 92.94 & 93.81 & \textbf{+0.86} \\
\addlinespace
\multirow{4}{*}{SemEval-2018} & GPT-5 {\scriptsize (CoT)} & 85.46 & 89.03 & \textbf{+3.57} & 85.99 & 89.16 & \textbf{+3.17} \\
 & GPT-5 {\scriptsize (IO)} & 85.37 & 86.73 & \textbf{+1.36} & 85.87 & 86.70 & \textbf{+0.83} \\
 & Qwen3-8B {\scriptsize (CoT)} & 66.48 & 78.19 & \textbf{+11.71} & 66.07 & 78.17 & \textbf{+12.09} \\
 & Qwen3-8B {\scriptsize (IO)} & 49.13 & 61.73 & \textcolor{red}{\textbf{+12.60}} & 44.02 & 60.65 & \textcolor{red}{\textbf{+16.63}} \\
\addlinespace
\multirow{4}{*}{iSarcasmEval} & GPT-5 {\scriptsize (CoT)} & 82.93 & 86.00 & \textbf{+3.07} & 75.25 & 78.18 & \textbf{+2.93} \\
 & GPT-5 {\scriptsize (IO)} & 82.79 & 83.86 & \textbf{+1.07} & 74.95 & 75.97 & \textbf{+1.02} \\
 & Qwen3-8B {\scriptsize (CoT)} & 58.57 & 72.21 & \textcolor{red}{\textbf{+13.64}} & 52.99 & 63.13 & \textbf{+10.14} \\
 & Qwen3-8B {\scriptsize (IO)} & 32.07 & 43.93 & \textbf{+11.86} & 31.97 & 42.30 & \textcolor{red}{\textbf{+10.33}} \\
\bottomrule
\end{tabular}
\end{table*}

\subsection{Correlation Between Reasoning Length and Performance}
\begin{figure}[t]
    \centering
    \begin{subfigure}[t]{0.48\textwidth}
        \centering
        \includegraphics[width=\linewidth]{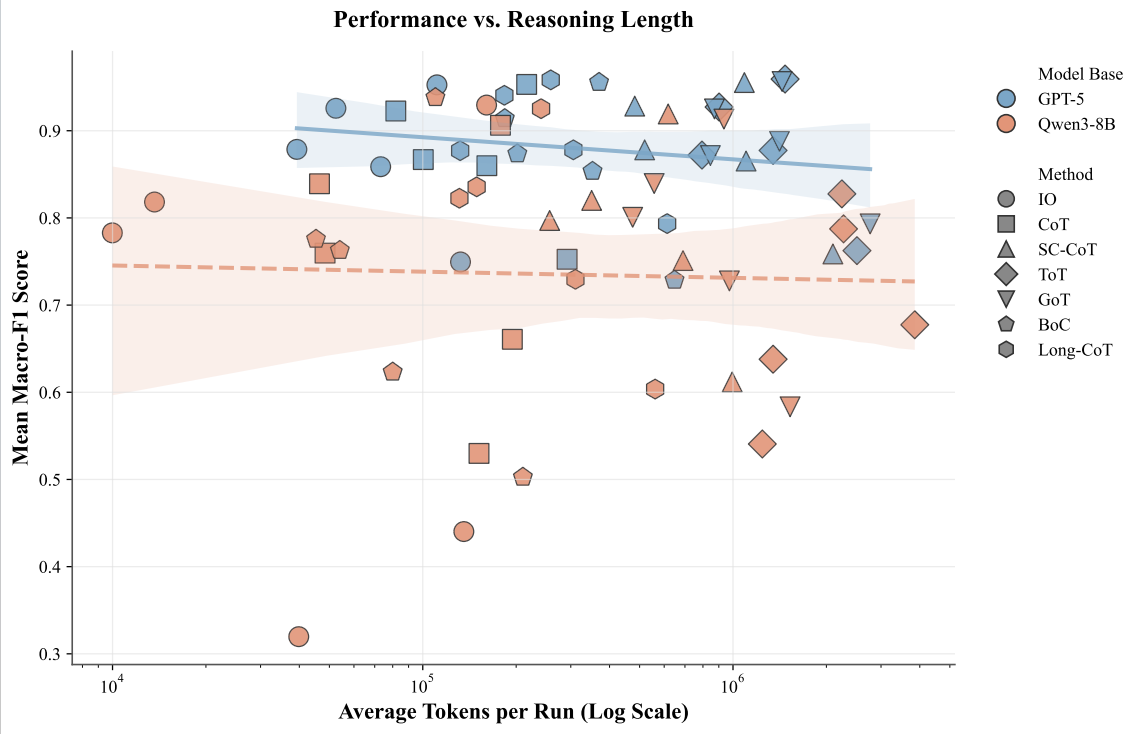}
        \caption{Average reasoning length across methods.}
    \end{subfigure}
    \hfill
    \begin{subfigure}[t]{0.46\textwidth}
        \centering
        \includegraphics[width=\linewidth]{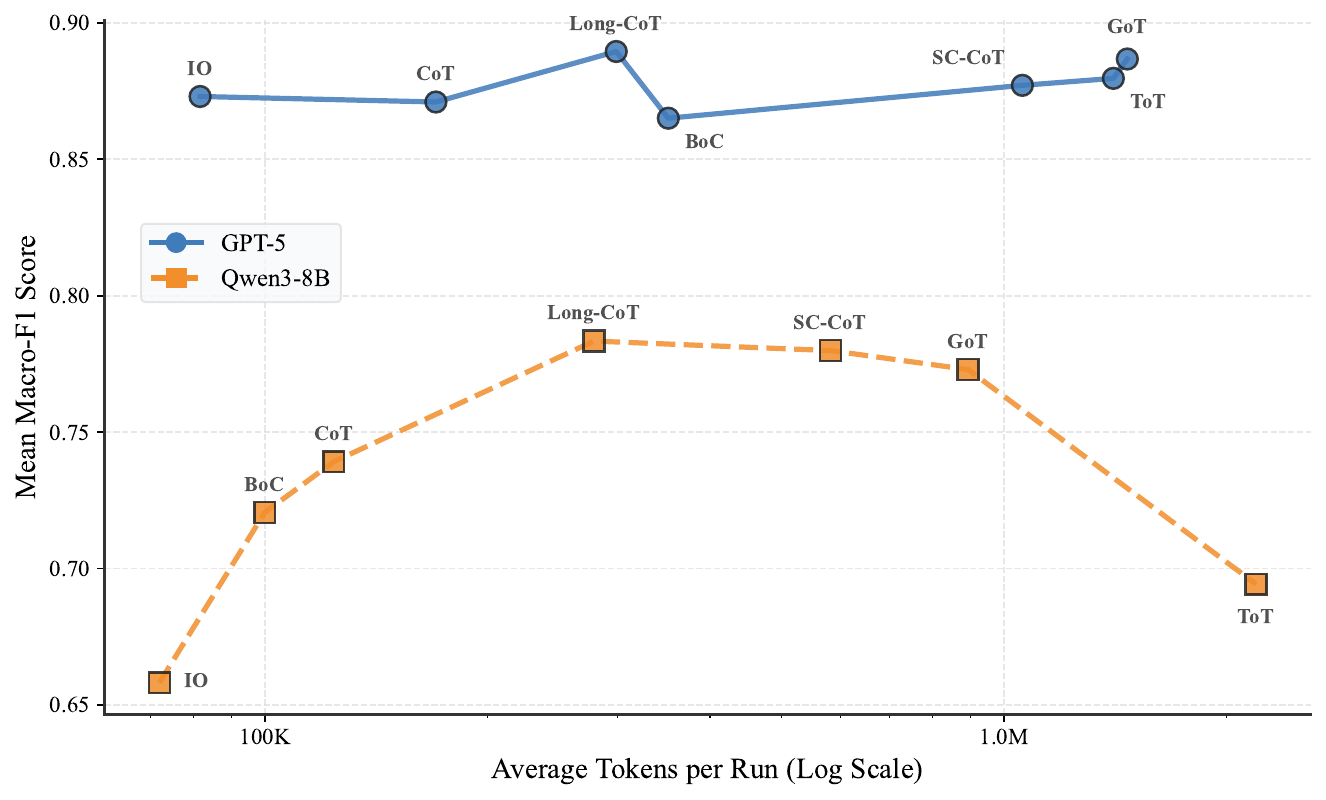}
        \caption{Performance vs. reasoning length.}
    \end{subfigure}
    \caption{Analysis of reasoning length and its relationship with performance.}
    \label{fig:length_performance}
\end{figure}

We analyze the relationship between reasoning length (measured by average tokens per run) and classification performance, as illustrated in Fig.~\ref{fig:length_performance}. 

For GPT-5, increasing reasoning length does not consistently improve performance. 
When moving from IO (81{,}667 tokens, 0.8730) to CoT (170{,}244 tokens, 0.8710), performance slightly decreases, indicating that moderate-length reasoning does not necessarily provide additional benefit. 
Further increases in length, such as SC-CoT (1{,}058{,}046 tokens, 0.8771) and ToT (1{,}404{,}805 tokens, 0.8797), yield only marginal gains. 
Notably, the best performance is achieved by Long-CoT (298{,}577 tokens, 0.8895) and GoT (1{,}467{,}153 tokens, 0.8868), suggesting that for big models, performance improvements depend more on the effectiveness of reasoning structure than on sheer length. 

In contrast, Qwen3-8B exhibits a typical inverted-U trend. 
Performance improves substantially as reasoning length increases from IO (71{,}938 tokens, 0.6581) to CoT (123{,}828 tokens, 0.7391) and SC-CoT (581{,}748 tokens, 0.7799). 
However, when the reasoning length becomes excessively large, as in ToT (2{,}191{,}753 tokens, 0.6942), performance drops significantly, falling below SC-CoT and GoT. 
This suggests that smaller models are more susceptible to overextended reasoning, where longer chains introduce noise, error propagation, and redundant expansions that degrade performance.

Overall, these results indicate that there is no stable positive correlation between reasoning length and performance. 
Insufficient length may fail to capture necessary semantic dependencies, while excessive length leads to overthinking, where additional tokens do not translate into meaningful performance gains.

\begin{table*}[t]
\centering
\caption{Representative examples from the three case types in our error analysis.}
\label{tab:error-analysis}
\footnotesize
\setlength{\tabcolsep}{4.5pt}
\renewcommand{\arraystretch}{1.35}
\setlength{\heavyrulewidth}{0.8pt}
\setlength{\lightrulewidth}{0.8pt}

\begin{tabularx}{\textwidth}{
    >{\centering\arraybackslash}m{0.8cm}
    >{\centering\arraybackslash}m{1.45cm}
    >{\raggedright\arraybackslash}X
    >{\centering\arraybackslash}m{1.55cm}
    >{\centering\arraybackslash}m{1.45cm}
    >{\centering\arraybackslash}m{1.55cm}
}
\toprule

\rowcolor{casecolor}
\multicolumn{6}{l}{\textbf{Case 1: Over-reasoning after a correct intuitive prediction}} \\
\midrule
ID & Method & Example & True Label & IO & Reasoning \\
\midrule

1 & ToT & \cellcolor{examplerow} Indian state rolls out wireless broadband. Government in South Indian state of Kerala sets up wireless kiosks as part of initiative to bridge digital divide. & Sci/Tech & Sci/Tech & \textcolor{errorred}{World} \\

2 & GoT & \cellcolor{examplerow} High court hears dispute over Michigan interstate wine sales. The Supreme Court is considering whether Michigan and other states may bar people from buying wine directly from out-of-state suppliers, a big-money question that could lead to sweeping changes in how alcoholic beverages are regulated. & Business & Business & \textcolor{errorred}{World} \\

3 & SC-CoT & \cellcolor{examplerow} What river in the US is known as the Big Muddy? & Location & Location & \textcolor{errorred}{Entity} \\

4 & CoT & \cellcolor{examplerow} Who was Abraham Lincoln? & Human & Human & \textcolor{errorred}{Description} \\

\midrule

\rowcolor{casecolor}
\multicolumn{6}{l}{\textbf{Case 2: Reasoning corrects an initially wrong intuitive judgment}} \\
\midrule
ID & Method & Example & True Label & IO & Reasoning \\
\midrule

5 & CoT & \cellcolor{examplerow} What is the deepest lake in the US? & Location & \textcolor{errorred}{Entity} & Location \\

6 & CoT & \cellcolor{examplerow} Dry your eyes mate, they're only trousers. & ironic & \textcolor{errorred}{non-ironic} & ironic \\

7 & CoT & \cellcolor{examplerow} (d)oes n't bother being as cloying or preachy as equivalent evangelical christian movies -- maybe the filmmakers know that the likely audience will already be among the faithful. & positive & \textcolor{errorred}{negative} & positive \\

8 & BoC & \cellcolor{examplerow} Broadcaster Donates \#36;325,000 to GOP (AP). One of the state's biggest broadcasters has given 13 Republican county committees \#36;325,000 worth of free air time to promote candidates on its radio and television stations throughout California. & World & \textcolor{errorred}{Business} & World \\

\midrule

\rowcolor{casecolor}
\multicolumn{6}{l}{\textbf{Case 3: Larger models benefit more reliably from reasoning}} \\
\midrule
ID & Method & Example & True Label & Qwen3-8B & GPT-5 \\
\midrule

9 & SC-CoT & \cellcolor{examplerow} One of the creepiest, scariest movies to come along in a long, long time, easily rivaling \textit{Blair Witch} or \textit{The Others}. & positive & \textcolor{errorred}{negative} & positive \\

10 & SC-CoT & \cellcolor{examplerow} IBM launches top-end Power5 servers. IBM has expanded the top end of its eServer range with three multiple-processor systems aimed at datacentres and large enterprise clients. & Sci/Tech & \textcolor{errorred}{Business} & Sci/Tech \\

11 & GoT & \cellcolor{examplerow} Who developed the Macintosh computer? & Human & \textcolor{errorred}{Entity} & Human \\

12 & ToT & \cellcolor{examplerow} On the train and surrounded by posh people, I'm so at home! \#not \#stickoutlikeasorethumb & ironic & \textcolor{errorred}{non-ironic} & ironic \\

\bottomrule
\end{tabularx}
\end{table*}

\subsection{Error Analysis}
To better understand when and why reasoning helps or harms performance, we conduct a qualitative error analysis based on representative cases in Table~\ref{tab:error-analysis}. The results reveal three consistent patterns.

First, \textbf{over-reasoning can degrade correct predictions}. In several examples, the IO baseline already identifies the correct label based on salient lexical cues, while more complex reasoning (e.g., ToT and GoT) introduces unnecessary semantic expansion and shifts attention toward less relevant concepts. This often leads to category drift, such as misclassifying \textit{Sci/Tech} or \textit{Business} as \textit{World}. 

Second, \textbf{reasoning is beneficial when the task requires implicit semantic or pragmatic inference}. In cases involving entity typing, sarcasm detection, or sentiment with implicit polarity, reasoning helps recover the correct label by making latent constraints explicit. For example, CoT can correct misclassifications from \textit{Entity} to \textit{Location}, or from \textit{non-ironic} to \textit{ironic}, where surface cues alone are insufficient. This explains why moderate reasoning methods (e.g., CoT and SC-CoT) achieve consistent improvements on more subjective tasks such as SemEval-2018 and iSarcasmEval.

Third, \textbf{the effectiveness of reasoning is strongly dependent on model capacity}. Comparing Qwen3-8B and GPT-5 on identical examples, we observe that bigger models are more capable of leveraging reasoning to refine predictions, while smaller models are more prone to distraction and error propagation. For instance, GPT-5.2 successfully resolves sentiment, topic, and entity classification through structured reasoning, whereas Qwen3-8B often remains sensitive to superficial or misleading cues. This observation is consistent with our broader results showing that reasoning benefits are more stable and reliable for big models.

Overall, these findings indicate that reasoning in text classification is not universally beneficial. Its effectiveness depends on whether it reinforces task-relevant evidence, whether the task requires deeper semantic inference, and whether the model has sufficient capacity to control the reasoning process without introducing noise or drift.

\section{Conclusion and Future Work}
In this paper, we conduct a systematic study of reasoning strategies for text classification across multiple models, tasks, and inference paradigms. We benchmark seven representative reasoning methods on both small and large language models, and introduce two efficiency metrics, Absolute Efficiency (AE) and Marginal Efficiency (ME), to quantify the trade-off between performance and computational cost. 
Our results show that reasoning does not consistently improve performance: simple methods such as IO and CoT are often competitive, while more complex strategies (e.g., ToT, GoT) frequently introduce higher cost and instability without proportional gains. 
Through the proposed efficiency metrics (AE and ME), we further demonstrate that longer or more complex reasoning is generally inefficient, and that performance is not positively correlated with reasoning length. 
Stability and few-shot analyses reinforce that effective reasoning depends on model capacity and task characteristics rather than reasoning complexity alone. 
Overall, our findings suggest that reasoning should be applied selectively and cost-aware, and future work should focus on developing adaptive and efficient reasoning mechanisms.

\begin{acks}
This paper is supported by The National Social Science Fund of China (No. 25AYY001) and a grant under the Collaborative Research with World-leading Research Groups scheme in The Hong Kong Polytechnic University (project no. G-SACF) and a General Research Fund under Hong Kong Research Grants Council (project no. 15218521). 
\end{acks}

\bibliographystyle{ACM-Reference-Format}
\bibliography{reference}










\end{document}